\documentclass[10pt,journal,compsoc]{IEEEtran}
\ifCLASSOPTIONcompsoc
  \usepackage[nocompress]{cite}
\else
  \usepackage{cite}
\fi
\ifCLASSINFOpdf
\else
\fi
\usepackage{caption}
\usepackage{subcaption}
\usepackage{graphicx}
\usepackage{amsmath,amssymb} % define this before the line numbering.
\usepackage{epsfig}
\usepackage{graphicx}
\usepackage{bbm}
\usepackage{bm}
\usepackage{amsmath}
\usepackage{amssymb}
\usepackage{url}
\usepackage{multirow}
\usepackage{siunitx}
\usepackage{amsfonts}
\usepackage{bbm}
\usepackage[export]{adjustbox}
\usepackage{enumitem}
\usepackage[space]{grffile}
\usepackage{enumitem}
\setlist{nolistsep}
\usepackage{color}

\usepackage{url}  %Required
\usepackage{graphicx}  %Required

\usepackage{url}            % simple URL typesetting
\usepackage{booktabs}       % professional-quality tables
\usepackage{amsfonts}       % blackboard math symbols
\usepackage{nicefrac}       % compact symbols for 1/2, etc.
\usepackage{microtype}      % microtypography
\usepackage{epsfig}
\usepackage{amsmath}
\usepackage{amssymb}
\usepackage{tabularx}
\usepackage{xargs} 
\usepackage{mathtools}
\usepackage{bm}
\usepackage{subcaption} 
\usepackage{enumitem}
\setlist{nolistsep}  % For reducing space between enum items (enumitem)
\usepackage{multirow}
\usepackage{pifont} 

\newcommand{\tickm}{\ding{52}}
\newcommand{\crossm}{\ding{53}}
\usepackage{xcolor,colortbl}
\definecolor{Gray}{gray}{0.85}
\definecolor{LightCyan}{rgb}{0.88,1,1}
\definecolor{LightPink}{rgb}{1, 0.8, 0.6}
\newcolumntype{Y}{>{\raggedleft\arraybackslash}X}
\newcolumntype{Z}{>{\columncolor{Gray}}Y}
\newcolumntype{U}{>{\columncolor{LightCyan}}Y}
\newcolumntype{V}{>{\columncolor{LightPink}}Y}

\usepackage[breaklinks=true,bookmarks=false,colorlinks=true]{hyperref}

\begin{document}
\title{Deep Recurrent Models of Pictionary-style Word Guessing}

\author{Ravi~Kiran~Sarvadevabhatla,~\IEEEmembership{Member,~IEEE,}
         Shiv~Surya,
         Trisha~Mittal~and~R.~Venkatesh~Babu~\IEEEmembership{Senior Member,~IEEE}
\thanks{Contact: \texttt{ravika@gmail.com}}}

\IEEEtitleabstractindextext{%
\begin{abstract}
The ability of intelligent agents to play games in human-like fashion is popularly considered a benchmark of progress in Artificial Intelligence. Similarly, performance on multi-disciplinary tasks such as Visual Question Answering (VQA) is considered a marker for gauging progress in Computer Vision. In our work, we bring games and VQA together. Specifically, we introduce the first computational model aimed at Pictionary, the popular word-guessing social game. We first introduce Sketch-QA, an elementary version of Visual Question Answering task. Styled after Pictionary, Sketch-QA uses incrementally accumulated sketch stroke sequences as visual data. Notably, Sketch-QA involves asking a  fixed question (``What object is being drawn?") and gathering open-ended guess-words from human guessers. We analyze the resulting dataset and present many interesting findings therein. To mimic Pictionary-style guessing, we subsequently propose a deep neural model which generates guess-words in response to temporally evolving human-drawn sketches. Our model even makes human-like mistakes while guessing, thus amplifying the human mimicry factor. We evaluate our model on the large-scale guess-word dataset generated via Sketch-QA task and compare with various baselines. We also conduct a Visual Turing Test to obtain human impressions of the guess-words generated by humans and our model. Experimental results demonstrate the promise of our approach for Pictionary and similarly themed games.
\end{abstract}

}

\maketitle

\IEEEdisplaynontitleabstractindextext

% \fi
\IEEEpeerreviewmaketitle

\IEEEraisesectionheading{\section{Introduction}\label{sec:sgintro}}

In the history of AI, computer-based modelling of human player games such as Backgammon, Chess and Go has been an important research area. The accomplishments of well-known game engines (e.g. TD-Gammon~\cite{tesauro1994td}, DeepBlue~\cite{DBLP:conf/aaai/1997w6}, AlphaGo~\cite{silver2016mastering}) and their ability to mimic human-like game moves has been a well-accepted proxy for gauging progress in AI. Meanwhile, progress in visuo-lingual problems such as visual captioning~\cite{chen2015mind,venugopalan2015sequence,xu2015show} and visual question answering~\cite{antol2015vqa,Xu2016,ren2015exploring} is increasingly serving a similar purpose for computer vision community. With these developments as backdrop, we explore the popular social game Pictionary\textsuperscript{TM}. 

The game of Pictionary brings together predominantly the visual and linguistic modalities. The game uses a shuffled deck of cards with guess-words printed on them. The participants first group themselves into teams and each team takes turns. For a given turn, a team's member selects a card. He/she then attempts to draw a sketch corresponding to the word printed on the card in such a way that the team-mates can guess the word correctly. The rules of the game forbid any verbal communication between the drawer and team-mates. Thus, the drawer conveys the intended guess-word primarily via the sketching process. 

\begin{figure}[!htbp]
    \centering
    \noindent
    \includegraphics[width=\linewidth]{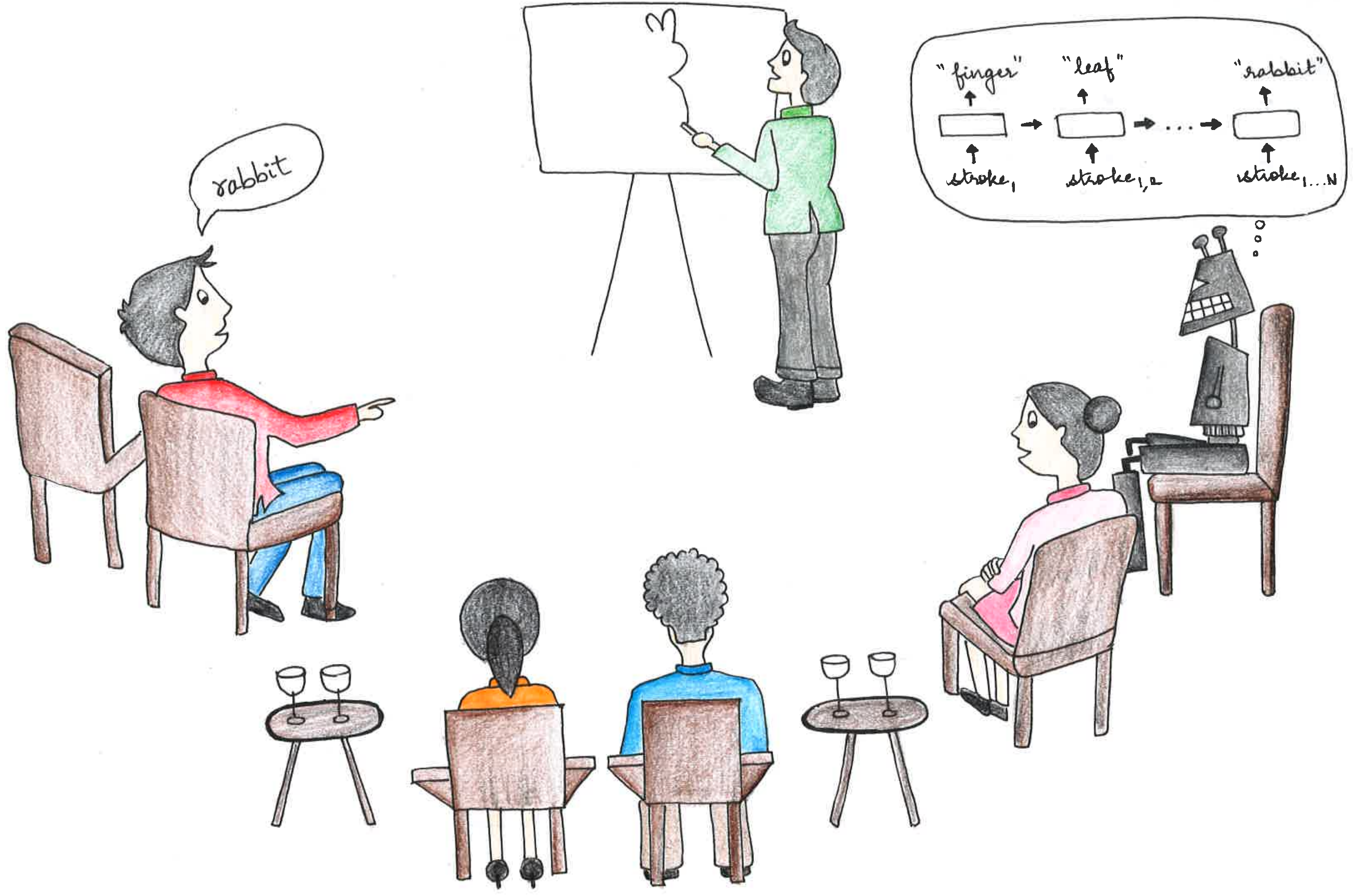}
    \caption{We propose a deep recurrent model of Pictionary-style word guessing. Such models can enable social robots to participate in real-life game scenarios as shown above. Picture credit:Trisha Mittal.}
    \label{fig:robotpictionary}
\end{figure}

Consider the scenario depicted in Figure \ref{fig:robotpictionary}. A group of people are playing Pictionary. New to the game, a  `social' robot is watching people play. Passively, its sensors record the strokes being drawn on the sketching board, guess-words uttered by the drawer's team members and finally, whether the last guess is correct. Having observed multiple such game rounds, the robot learns computational models which mimic human guesses and enable it to participate in the game.  

\begin{figure*}[!ht]
    \centering
    \noindent
    \includegraphics[width=\linewidth]{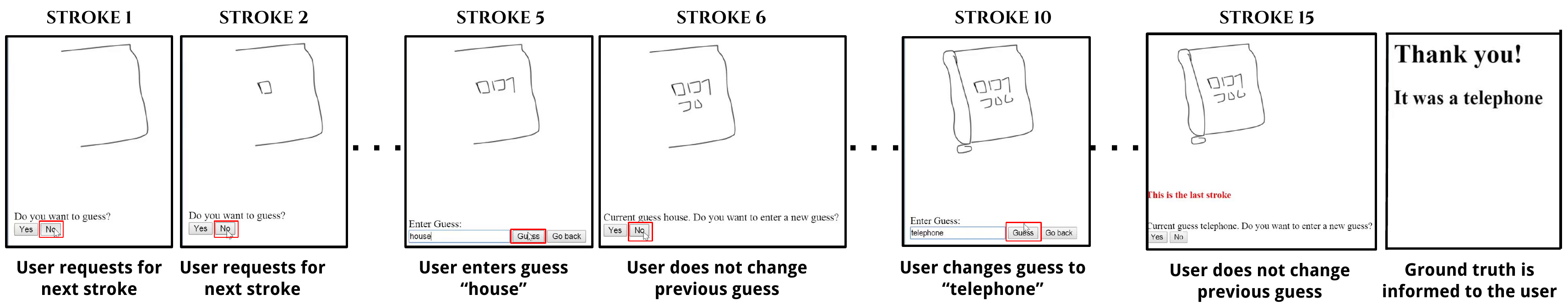}
    \caption{The time-line of a typical Sketch-QA guessing session: Every time a stroke is added, the subject either inputs a best-guess word of the object being drawn (stroke \#$5,10$). In case existing strokes do not offer enough clues, he/she requests the next stroke be drawn. After the final stroke (\#$15$), the subject is informed the object's ground-truth category.}
    \label{fig:overview}
\end{figure*}

As a step towards building such computational models, we first collect guess-word data via Sketch Question Answering (Sketch-QA), a novel, Pictionary-style guessing task. We employ a large-scale crowdsourced dataset of hand-drawn object sketches whose temporal stroke information is available~\cite{eitz2012humans}. Starting with a blank canvas, we successively add strokes of an object sketch and display this process to human subjects (see Figure \ref{fig:overview}). Every time a stroke is added, the subject provides a best-guess of the object being drawn. In case existing strokes do not offer enough clues for a confident guess, the subject requests the next stroke be drawn. After the final stroke, the subject is informed the object category. 

Sketch-QA can be viewed as a rudimentary yet novel form of Visual Question Answering (VQA)~\cite{antol2015vqa,ren2015exploring,Xu2016,venugopalan2015sequence}. Our approach differs from existing VQA work in that [a] the visual content consists of sparsely detailed hand-drawn depictions  [b] the visual content necessarily accumulates over time [c] at all times, we have the same question -- ``What is the object being drawn?" [d] the answers (guess-words) are open-ended (i.e. not 1-of-K choices) [e] for a while, until sufficient sketch strokes accumulate, there may not be `an answer'. Asking the same question might seem an oversimplification of VQA. However, other factors --- extremely sparse visual detail, inaccuracies in object depiction arising from varying drawing skills of humans and open-ended nature of answers --- pose unique challenges that need to be addressed in order to build viable computational models. 

Concretely, we make the following contributions:
 \begin{itemize}
     \item  We introduce a novel task called Sketch-QA to serve as a proxy for Pictionary (Section \ref{sec:sgmethodology}).
     \item  Via Sketch-QA, we create a new crowdsourced dataset of paired guess-word and sketch-strokes, dubbed \textsc{WordGuess-160}, collected from $16{,}624$ guess sequences of $1{,}108$ subjects across $160$ sketch object categories.      
     \item We perform comparative analysis of human guessers and a machine-based sketch classifier via the task of sketch recognition (Section \ref{sec:finalguessword}).
     \item We introduce a novel computational model for word guessing (Section \ref{sec:sgcompmodels}). Using \textsc{WordGuess-160} data, we analyze the performance of the model for Pictionary-style on-line guessing and conduct a Visual Turing Test to gather human assessments of generated guess-words (Section \ref{sec:sgDeepGuesser-eval}).
\end{itemize}

Please visit \url{github.com/val-iisc/sketchguess} for code and dataset related to this work. To begin with, we shall look at the procedural details involved in the creation of \textsc{WordGuess-160} dataset.

\section{Creating the \textsc{WordGuess-160} dataset}
\label{sec:sgdatacollection}

\subsection{Sketch object dataset} 
\label{sec:sgsketchobjdataset}

As a starting point, we use hand-sketched line drawings of single objects from the large-scale TU-Berlin sketch dataset~\cite{eitz2012humans}. This dataset contains $20{,}000$ sketches uniformly spread across $250$ object categories (i.e. $80$ sketches per category). The sketches were obtained in a crowd-sourced manner by providing only the category name (e.g. ``sheep'') to the sketchers. In this aspect, the dataset collection procedure used for TU-Berlin dataset aligns with the draw-using-guess-word-only paradigm of Pictionary. For each sketch object,  temporal order in which the strokes were drawn is also available. A subsequent analysis of the TU-Berlin dataset by Schneider and Tuytelaars~\cite{Schneider:2014:SCC:2661229.2661231} led to the creation of a curated subset of sketches which were deemed visually less ambiguous by human subjects. For our experiments, we use this curated dataset containing $160$ object categories with an average of $56$ sketches per category. 

\subsection{Data collection methodology} 
\label{sec:sgmethodology}

To collect guess-word data for Sketch-QA, we used a web-accessible crowdsourcing portal. Registered participants were initially shown a screen displaying the first stroke of a randomly selected sketch object from a randomly chosen category (see Figure \ref{fig:overview}). A GUI menu with options `Yes',`No' was provided. If the participants felt more strokes were needed for guessing, they clicked the `No' button, causing the next stroke to be added. On the other hand, clicking `Yes' would allow them to type their current best guess of the object category. If they wished to retain their current guess, they would click `No', causing the next stroke to be added. This act (clicking `No') also propagates the most recently typed guess-word and associates it with the strokes accumulated so far. The participant was instructed to provide guesses as early as possible and as frequently as required. After the last stroke is added, the ground-truth category was revealed to the participant. Each participant was encouraged to guess a minimum of $125$ object sketches. Overall, we obtained guess data from $1{,}108$ participants. 

Given the relatively unconstrained nature of guessing, we pre-process the guess-words to eliminate artifacts as described below.

\subsection{Pre-processing}
\label{sec:sgpreprocessing}

\textbf{Incomplete Guesses:} In some instances, subjects provided guess attempts for initial strokes but entered blank guesses subsequently. For these instances, we propagated the last non-blank guess until the end of stroke sequence.

\textbf{Multi-word Guesses:} In some cases, subjects provided multi-word phrase-like guesses (e.g. ``pot of gold at the end of the rainbow" for a sketch depicting the object category \texttt{rainbow}). Such guesses seem to be triggered by extraneous elements depicted in addition to the target object. For these instances, we used the HunPos tagger~\cite{halacsy2007hunpos} to retain only the noun word(s) in the phrase.

\textbf{Misspelt Guesswords:} To address incorrect spellings, we used the Enchant spellcheck library~\cite{enchant} with its default \textit{Words} set augmented with the $160$ object category names from our base dataset~\cite{eitz2012humans} as the spellcheck dictionary.

\textbf{Uppercase Guesses:} In some cases, the guess-words exhibit non-uniform case formatting (e.g. all uppercase or a mix of both uppercase and lowercase letters). For uniformity, we formatted all words to be in lowercase.

In addition, we manually checked all of the guess-word data to remove unintelligible and inappropriate words. We also removed sequences that did not contain any guesses. Thus, we finally obtain the \textsc{GuessWord-160} dataset comprising of guesswords distributed across $16{,}624$ guess sequences and $160$ categories. It is important to note that the final or the intermediate guesses could be `wrong', either due to the quality of drawing or due to human error. We deliberately do not filter out such guesses. This design choice keeps our data realistic and ensures that our computational model has the opportunity to characterize both the `success' and `failure' scenarios of Pictionary.  

A video of a typical Sketch-QA session can be viewed at \url{https://www.youtube.com/watch?v=YU3entFwhV4}.

In the next section, we shall present various interesting facets of our \textsc{WordGuess-160} dataset.

\section{Guess Sequence Analysis}
\label{sec:guesseqanalysis}

\begin{figure}[!tbp]
    \centering
    \noindent
    \includegraphics[width=\linewidth]{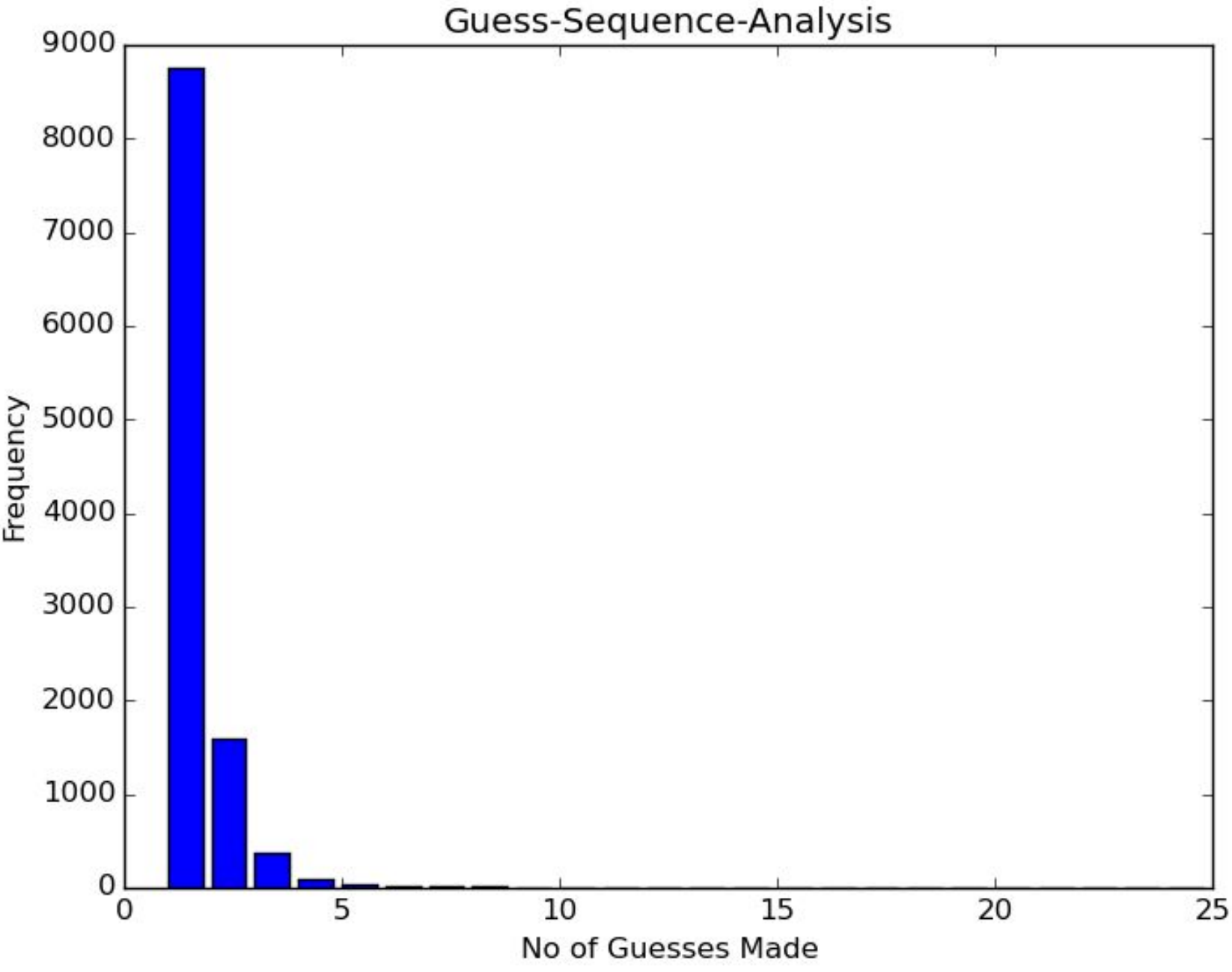}
    \caption{In the above plot, x-axis denotes the number of unique guesses. y-axis denotes the number of subjects who made corresponding number of unique guesses.}
    \label{fig:guesscount}
\end{figure}

Given a sketch, how many guesses are typically provided by subjects? To answer this, we examine the distribution of unique guesses per sequence. As Figure \ref{fig:guesscount} shows, the number of guesses have a large range. This is to be expected given the large number of object categories we consider and associated diversity in depictions. A large number of subjects provide a single guess. This arises both from the inherent ambiguity of the partially rendered sketches and the confidence subjects place on their guess. This observation is also borne out by Table \ref{tab:appgh} which shows the number of sequences eliciting $x$ guesses ($x = \{1,2,3,\geqslant 4\}$).

\begin{table}[h]
\centering
\begin{tabular}{ c || c c c c}
\hline
Guesses & $1$ & $2$ & $3$ & $\geqslant 4$ \\
\hline
\# Sequences & $12520$ & $2643$ & $568$ & $279$ \\
\hline
\end{tabular}
\caption{The distribution of possible number of guesses and count of number of sequences which elicited them.}
\label{tab:appgh}
\end{table}

\begin{figure}[!ht]
    \centering
    \noindent
    \includegraphics[width=\linewidth]{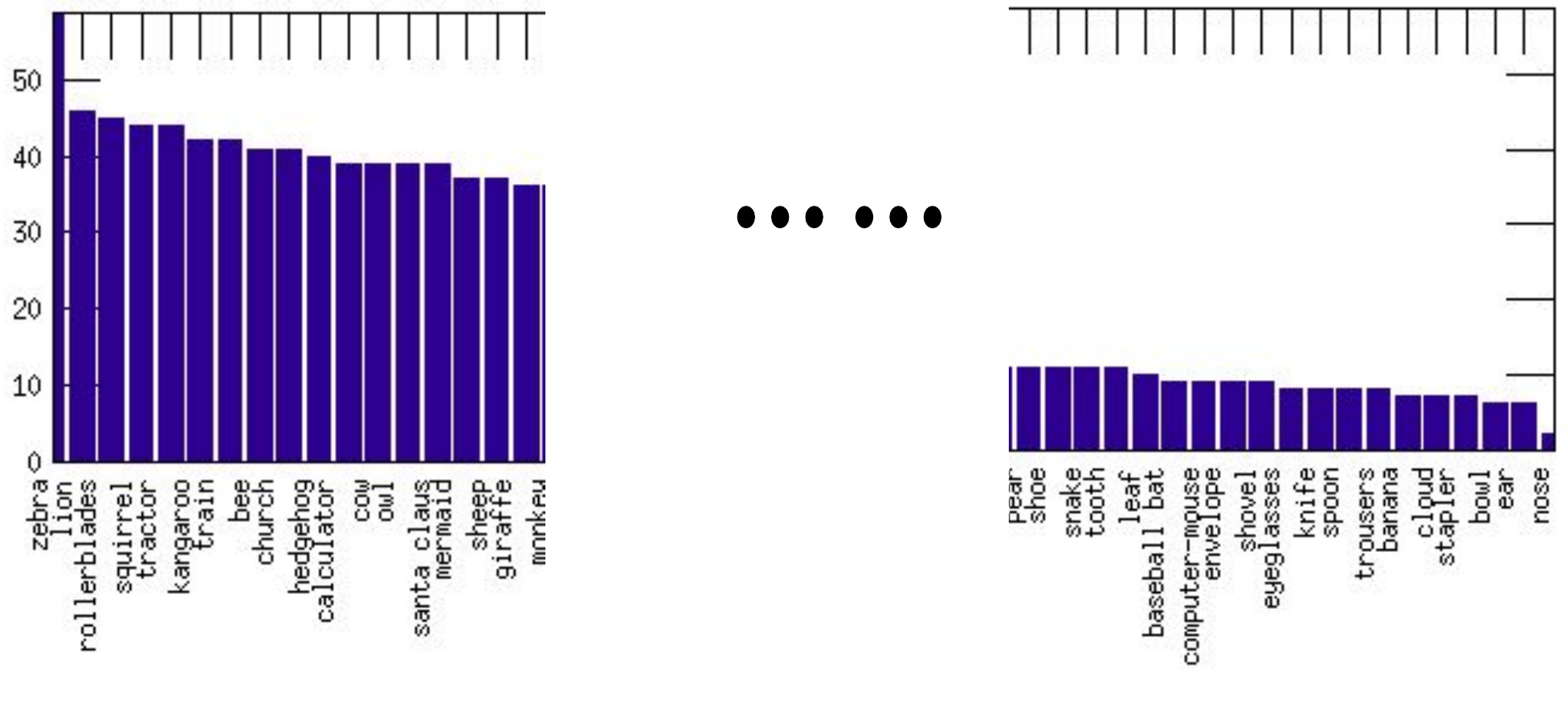}
    \caption{Here, x-axis denotes the categories. y-axis denotes the number of sketches within the category with multiple guesses. The categories are shown sorted by the number of sketches which elicited multiple guesses.}
    \label{fig:appguesscountcat}
\end{figure}

We also examined the sequences which elicited multiple guesses in terms of object categories they belong to. The categories were sorted by the number of multi-guess sequences their sketches elicited. The top-$10$ and bottom-$10$ categories according to this criteria can be viewed in Figure \ref{fig:appguesscountcat}. This perspective helps us understand which categories are inherently ambiguous in terms of their stroke-level evolution when usually drawn by humans.

\begin{figure*}[!t]
\centering
\includegraphics[width=\linewidth]{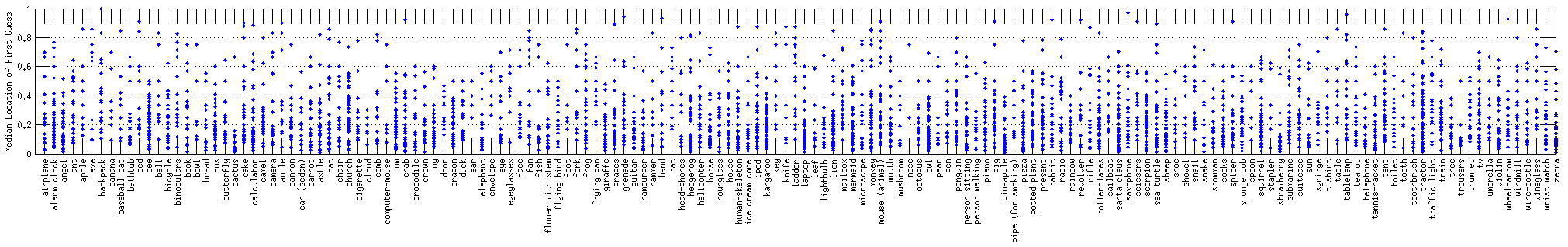}
\caption{The distribution of first guess locations normalized $([0,1])$ over sequence lengths (y-axis) across categories (x-axis). }
\label{fig:firstguessloc}
\end{figure*}

Another interesting statistic is the distribution of first guess location relative to length of the sequence. Figure \ref{fig:firstguessloc} shows the distribution of first guess index locations as a function of sequence length (normalized to $1$). Thus, a value closer to $1$ implies that the first guess was made late in the sketch sequence. Clearly, the guess location has a large range across the object categories. The requirement to accurately capture this range poses a considerable challenge for computational models of human guessing. 

\begin{figure}[!ht]
\centering
\includegraphics[width=\linewidth]{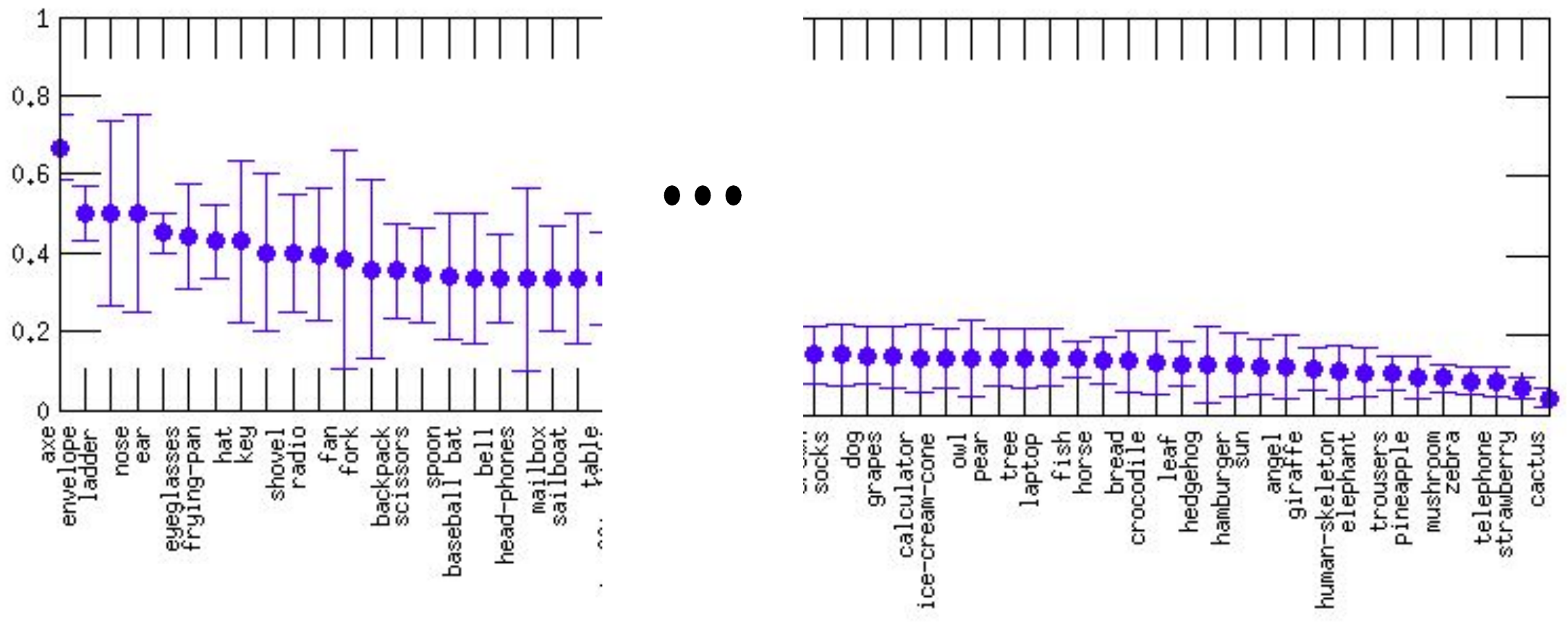}
\caption{Categories sorted by the median location of first guess. }
\label{fig:appfirstguessloccat}
\end{figure}

To obtain a category-level perspective, we computed the median first-guess location and corresponding deviation of first guess location on a per-category basis and sorted the categories by the median values. The resulting plot for the top and bottom categories can be viewed in Figure \ref{fig:appfirstguessloccat}. This perspective helps understand which the level at which categories evolve to a recognizable iconic stroke composition relative to the original, full-stroke reference sketch. Thus, categories such as \texttt{axe,envelope,ladder}, although seemingly simple, are depicted in a manner which induces doubt in the guesser, consequently delaying the induction of first guess. On the other hand, categories such as \texttt{cactus,strawberry,telephone} tend to be drawn such that the early, initial strokes capture the iconic nature of either the underlying ground-truth category or an easily recognizable object form different from ground-truth.

The above analysis focused mostly on the overall sequence-level trends in the dataset. In the next section, we focus on the last guess for each sketch stroke sequence. Since the final guess is associated with the full sketch, it can be considered the guesser's prediction of the object underlying the sketch. Such predictions can then be compared with ground-truth labels originally provided with the sketch dataset to determine `human guesser' accuracy  (Section \ref{sec:appclassifperf}). Subsequently, we compare `human guesser' accuracy with that of a machine-based sketch object recognition classifier and discuss trends therein  (Section \ref{sec:appcomparehumanmachine}).

\section{Final guess-word analysis}
\label{sec:finalguessword}

\begin{table*}[!h]
\resizebox{\textwidth}{!}
{
\centering
\begin{tabular}{c|cccccc}
\toprule
\scriptsize{Criteria Combination} & \scriptsize{EM} &  \scriptsize{EM $\mid$ SUB} & \scriptsize{EM $\mid$ SUB $\mid$ SYN} & \scriptsize{EM $\mid$ SUB $\mid$ SYN $\mid$ HY} & \scriptsize{EM $\mid$ SUB $\mid$ SYN $\mid$ HY $\mid$ HY-PC} & \scriptsize{EM $\mid$ SUB $\mid$ SYN $\mid$ HY $\mid$ HY-PC $\mid$ WUP} \\
\midrule 
 \scriptsize{Accuracy} & $67.27$ & $75.49$ & $77.97$ & $80.08$ & $82.09$ & $83.33$ \\
\bottomrule
\end{tabular}
}
\caption{Accuracy of human guesses for various matching criteria (Section \ref{sec:appsimilaritymeasures}). The $\mid$ indicates that the matching criteria are combined in a logical-OR fashion to determine whether the predicted guess-word matches the ground-truth or not.}
\label{tab:appacc} 
\end{table*}

With \textsc{GuessWord-160} data at hand, the first question that naturally arises is: What is the ``accuracy" of humans on the final, full sketches (i.e. when all the original strokes have been included)? For a machine-based classifier, this question has a straightforward answer: Compute the fraction of sketches whose predicted category label is exactly the same as ground-truth. However, given the open-ended nature of guess-words, an `exact matching' approach is not feasible. Even assuming the presence of a universal dictionary, such an  approach is too brittle and restrictive. Therefore, we first define a series of semantic similarity criteria which progressively relax the correct classification criterion for the final sketches. 

\subsection{Matching criteria for correct classification}
\label{sec:appsimilaritymeasures}

\textbf{Exact Match (EM):} The predicted guess-word is a literal match (letter-for-letter) with the ground-truth category.

\noindent \textbf{Subset (SUB):} The predicted guess-word is a subset of ground-truth or vice-versa. This criteria lets us characterize certain multi-word guesses as correct (e.g. guess: \textit{pot of gold at the end of the rainbow}, ground-truth: \textit{rainbow}).

\noindent \textbf{Synonyms (SYN):} The predicted guess-word is a synonym of ground-truth. For synonym determination, we use the WordNet~\cite{miller1995wordnet} synsets of prediction and ground-truth. 

\noindent \textbf{Hypernyms (HY):} The one-level up parents (hypernyms) of ground-truth and predicted guess-word are the same in the hierarchy induced by WordNet graph. 

\noindent \textbf{Hypernyms-Parent and Child (HY-PC):} The ground-truth and prediction have a parent-child (hypernym) relationship in the WordNet graph. 

\noindent \textbf{Wu-Palmer Similarity (WUP)~\cite{wu1994verbs}:} This calculates relatedness of two words using a graph-distance based method applied to the corresponding WordNet synsets. If WUP similarity between prediction and ground-truth is at least $0.9$, we deem it a correct classification. 

\subsection{Classification Performance}
\label{sec:appclassifperf}

To compute the average accuracy of human guesses, we progressively relax the `correct classification' rule by combining the matching criteria (Section \ref{sec:appsimilaritymeasures}) in a logical-OR fashion. The average accuracy of human guesses can be viewed in Table \ref{tab:appacc}. The accuracy increases depending on the extent to which each successive criterion relaxes the base `exact match' rule. The large increase in accuracy for `EM $\mid$ SUB' (2nd row of the table) shows the pitfall of naively using the exact matching (1-hot label, fixed dictionary) rule.

\begin{table*}[!h]
\resizebox{\textwidth}{!}
{
\centering
\begin{tabular}{c|ccccc}
\toprule
\scriptsize{Criteria Combination} &  \scriptsize{EM $\mid$ SUB} & \scriptsize{EM $\mid$ SUB $\mid$ SYN} & \scriptsize{EM $\mid$ SUB $\mid$ SYN $\mid$ HY} & \scriptsize{EM $\mid$ SUB $\mid$ SYN $\mid$ HY $\mid$ HY-PC} & \scriptsize{EM $\mid$ SUB $\mid$ SYN $\mid$ HY $\mid$ HY-PC $\mid$ WUP} \\
\midrule 
 \scriptsize{Avg. rating} & $1.01$ & $\mathbf{1.93}$ & $0.95$ & $1.1$ & $0.21$ \\
\bottomrule
\end{tabular}
}
\caption{Quantifying the suitability of matching criteria combination for characterizing human-level sketch object recognition accuracy. The larger the human rating score, more suitable the criteria. See Section \ref{sec:appclassifperf} for details.}
\label{tab:appcriteria-ratings}
\end{table*}

At this stage, a new question arises: which of these criteria best characterizes human-level accuracy? Ultimately, ground-truth label is a consensus agreement among humans. To obtain such consensus-driven ground-truth, we performed a human agreement study. We displayed ``correctly classified'' sketches (w.r.t a fixed criteria combination from Table \ref{tab:appacc}) along with their labels, to human subjects. Note that the labelling chosen changes according to criteria combination. (e.g. A sketch with ground-truth \texttt{revolver} could be shown with the label \texttt{firearm} since such a prediction would be considered correct under the `EM $\mid$ SUB $\mid$ SYN $\mid$ HY' combination). Also, the human subjects weren't informed about the usage of criteria combination for labelling. Instead, they were told that the labellings were provided by other humans. Each subject was asked to provide their assessment of the labelling on a scale of $-2$ (`Strongly Disagree with labelling') to $2$ (`Strongly Agree with labelling'). We randomly chose $200$ sketches correctly classified under each criteria combination. For each sketch, we collected $5$ agreement ratings and computed the weighted average of the agreement score. Finally, we computed the average of these weighted scores. The ratings (Table \ref{tab:appcriteria-ratings}) indicate that `EM $\mid$ SUB $\mid$ SYN' is the criteria combination most agreed upon by human subjects for characterizing human-level accuracy. Having determined the criteria for a correct match, we can also contrast human-classification performance with a machine-based state-of-the-art sketch classifier. 

\section{Comparing human classification performance with a machine-based classifier}
\label{sec:appcomparehumanmachine}

 We contrast the human-level performance (`EM $\mid$ SUB $\mid$ SYN' criteria) with a state-of-the-art sketch classifier~\cite{Sarvadevabhatla:2016:EMR:2964284.2967220}. To ensure fair comparison, we consider only the $1204$ sketches which overlap with the test set used to evaluate the machine classifier. Table \ref{tab:app1} summarizes the prediction combinations (e.g. Human classification is correct, Machine classification is incorrect) between the classifiers. While the results seem to suggest that machine classifier `wins' over human classifier, the underlying reason is the open-ended nature of human guesses and the closed-world setting in which the machine classifier has been trained. 
 
To determine whether the difference between human and machine classifiers is statistically significant, we use the Cohen's $d$ test. Essentially, Cohen's $d$ is an effect size used to indicate the standardised difference between two means and ranges between $0$ and $1$. Suppose, for a given category $c$, the mean accuracy w.r.t human classification criteria is $\mu_h^c$ and the corresponding variance is $V_h^c$. Similarly, let the corresponding quantities for the machine classifier be $\mu_m^c$ and $V_m^c$. Cohen's $d$ for category $c$ is calculated as :
 
 \begin{align}
     d_c = \frac{\mu_m^c - \mu_h^c}{s}
 \end{align}
 
 where $s$ is the pooled standard deviation, defined as:
 
 \begin{align}
     s = \displaystyle \sqrt{\frac{V_m^c + V_h^c}{2}}
 \end{align}
 
We calculated Cohen's $d$ for all categories as indicated above and computed the average of resulting scores. The average value is $0.57$ which indicates significant differences in the classifiers according to the signficance reference tables commonly used to determine Cohen's $d$ significance. In general, though, there are categories where one classifier outperforms the other. The list of top-10 categories where one classifier outperforms the other (in terms of Cohen's $d$) is given in Table \ref{tab:app3}. 

\begin{table}[!h]
\centering
\footnotesize
\begin{tabular}{ c | c }
\hline
Machines outperform humans & Humans outperform machines \\
\hline 
scorpion ($0.84$) & dragon ($0.79$) \\
rollerblades ($0.82$) & owl ($0.75$) \\
person walking ($0.82$) & mouse ($0.72$) \\
revolver ($0.81$) & horse  ($0.72$) \\
sponge bob  ($0.81$) & flower with stem  ($0.71$) \\
rainbow  ($0.80$) & wine-bottle  ($0.65$) \\
person sitting  ($0.79$) & lightbulb  ($0.65$) \\
sailboat  ($0.79$) & snake ($0.63$) \\
suitcase  ($0.75$)  & leaf ($0.63$) \\
\hline
\end{tabular}
\caption{Category level performance of human and machine classifiers. The numbers alongside category names correspond to Cohen's $d$ scores.}
\label{tab:app3}
\end{table}

\begin{table}[!h]
\centering
\begin{tabular}{ c c c }
\hline
\multicolumn{2}{ c }{Prediction} & Relative \% of test data \\
\hline 
Human & Machine &  \\ \hline
\tickm & \crossm & $9.05$ \\ \hline
\crossm & \tickm & $20.43$ \\ \hline
\tickm & \tickm & $67.61$ \\ \hline
\crossm & \crossm & $2.91$ \\ \hline
\end{tabular}
\caption{Comparing human and machine classifiers for the possible prediction combinations -- \tickm \hspace{1mm} indicates correct and \crossm \hspace{1mm} indicates incorrect prediction.}
\label{tab:app1}
\end{table}

\begin{figure*}[!h]
    \centering
    \noindent
    \includegraphics[width=\linewidth]{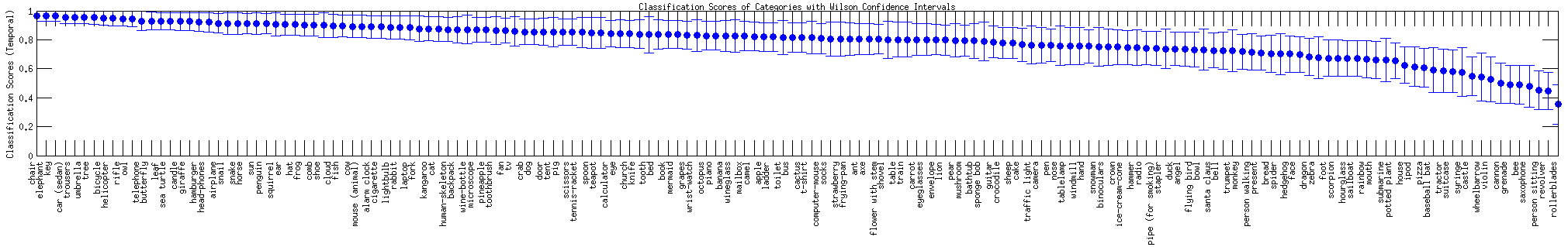}
    \caption{Distribution of correct predictions across categories, sorted by median category-level score. x-axis shows categories and y-axis stands for classification rate.}
    \label{fig:apppred-temporal-160}
\end{figure*}

The distribution of correct human guess statistics on a per-category basis can be viewed in Figure \ref{fig:apppred-temporal-160}. For each category, we calculate confidence intervals. These intervals inform us at a given level of certainty whether the true accuracy results will likely fall in the range identified. In particular, the Wilson score method of calculating confidence intervals, which we employ, assume that the variable of interest (the number of successes) can be modeled as a binomial random variable. Given that the binomial distribution can be considered the sum of $n$ Bernoulli trials, it is appropriate for our task, as a sketch is either classified correctly (success) or misclassified (failure). 

\begin{figure}[!h]
    \centering
    \noindent
    \includegraphics[width=\linewidth]{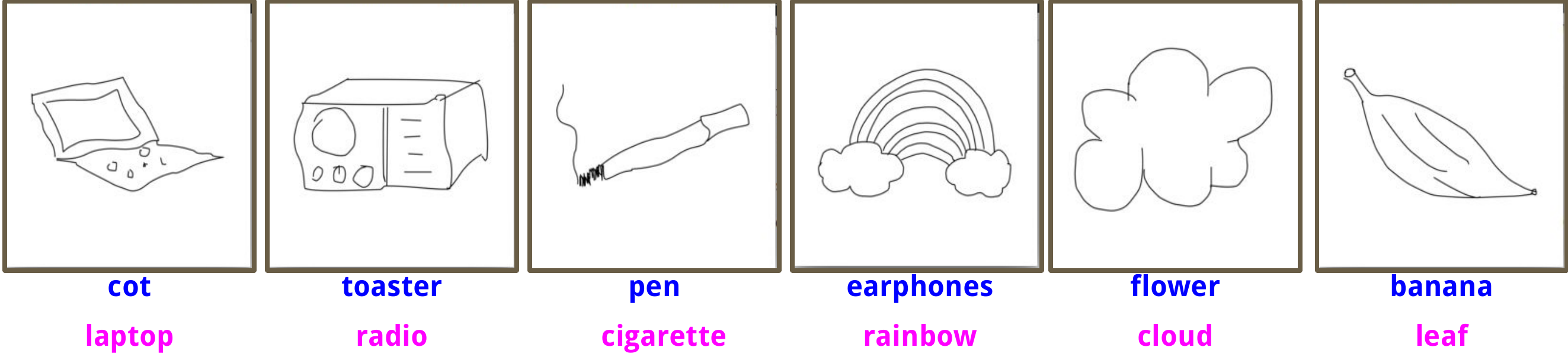}
    \caption{Some examples of misclassifications: Human guesses are shown in blue. Ground-truth category labels are in pink.}
    \label{fig:app2}
\end{figure}

Some examples of misclassifications (and the ground-truth category labels) can be seen in Figure \ref{fig:app2}. Although the guesses and ground-truth categories are lexically distant, the guesses are sensible when conditioned on visual stroke data.

\section{Computational Models}
\label{sec:sgcompmodels}

We now describe our computational model designed to produce human-like guess-word sequences in an on-line manner. For model evaluation, we split the $16624$ sequences in \textsc{GuessWord-160} randomly into disjoint sets containing $60\%$ , $25\%$ and $15\%$ of the data which are used during training, validation and testing phases respectively. 

\noindent \textbf{Data preparation:} Suppose a sketch $I$ is composed of $N$ strokes. Let the cumulative stroke sequence of $I$ be $\mathbb{I} = \{S_1,S_2,\ldots S_N\}$, i.e. $S_N=I$ (see Figure \ref{fig:overview}). Let the sequence of corresponding guess-words be $\mathbb{G_I} = \{g_1,g_2,\ldots g_N\}$. The sketches are first resized to $224 \times 224$ and zero-centered. To ensure sufficient training data, we augment sketch data and associated guess-words. For sketches, each accumulated stroke sequence $S_t \in \mathbb{I}$ is first morphologically dilated (`thickened'). Subsequent augmentations are obtained by applying vertical flip and scaling (paired combinations of $-7\%,-3\%,3\%,7\%$ scaling of image side). We also augment guess-words by replacing each guess-word in $\mathbb{G_I}$ with its plural form (e.g. \texttt{pant} is replaced by \texttt{pants}) and synonyms wherever appropriate. 

\noindent \textbf{Data representation:} The penultimate fully-connected layer's outputs of CNNs fine-tuned on sketches are used to represent sketch stroke sequence images. The guess-words are represented using pre-trained word-embeddings. Typically, a human-generated guess sequence contains two distinct phases. In the first phase, no guesses are provided by the subject since the accumulated strokes provide insufficient evidence. Therefore, many of the initial guesses ($g_1,g_2$ etc.) are empty and hence, no corresponding embeddings exist. To tackle this, we map `no guess' to a pre-defined non-word-embedding (symbol ``\#''). 

\begin{figure*}[!t]
\centering
\includegraphics[width=\textwidth,keepaspectratio]{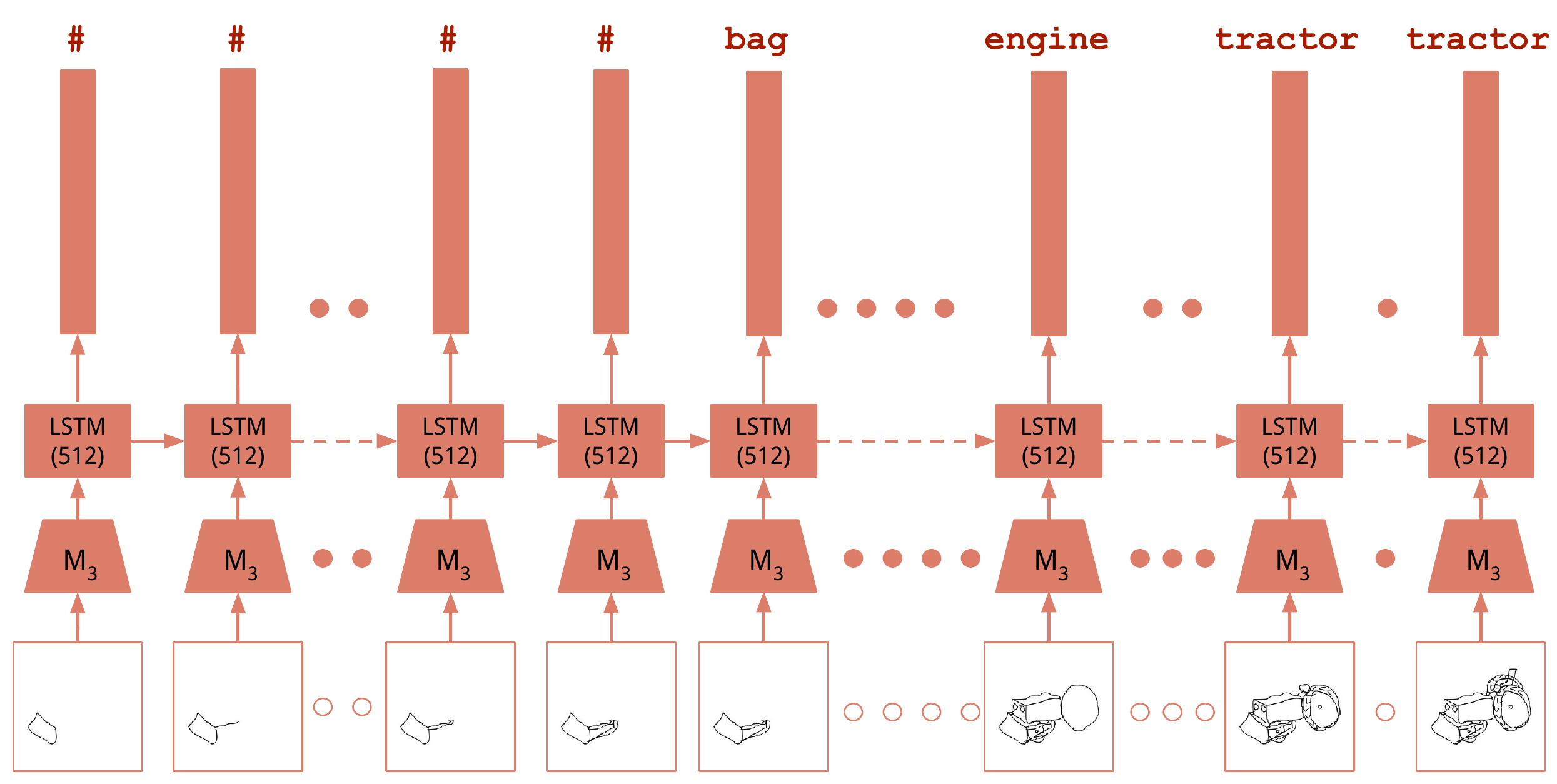}
\caption{The architecture for our deep neural model of word guessing. The rectangular bars correspond to guess-word embeddings. $M_3$ corresponds to the CNN regressor whose penultimate layer's outputs are used as input features to the LSTM model. ``\#'' reflects our choice of modelling `no guess' as a pre-defined non-word embedding. See Section \ref{sec:sgcompmodels} for details.}
\label{fig:DeepGuesser}
\end{figure*}

\noindent \textbf{Model design strategy:} Our model's objective is to map the cumulative stroke sequence $\mathbb{I}$ to a target guess-word sequence $\mathbb{G_I}$. Given our choice of data representation above, the model effectively needs to map the sequence of sketch features to a sequence of word-embeddings. To achieve this sequence-to-sequence mapping, we use a deep recurrent neural network (RNN) as the architectural template of choice (see Figure \ref{fig:DeepGuesser}). 

For the sequential mapping process to be effective, we need discriminative sketch representations. This ensures that the RNN can focus on modelling crucial sequential aspects such as when to initiate the word-guessing process and when to transition to a new guess-word once the guessing has begun (Section \ref{sec:sgfinetuning2}). To obtain discriminative sketch representations, we first train a CNN regressor to predict a guess-word embedding when an accumulated stroke image is presented  (Section \ref{sec:sgfinetuning}). It is important to note that we ignore the sequential nature of training data in the process. Additionally, we omit the sequence elements corresponding to `no-guess' during regressor training and evaluation. This frees the regressor from having to additionally model the complex many-to-one mapping between strokes accumulated before the first guess and a `no-guess'. 

To arrive at the final CNN regressor, we begin by fine-tuning a pre-trained photo object CNN. To minimize the impact of the drastic change in domain (photos to sketches) and task (classification to word-embedding regression), we undertake a series of successive fine-tuning steps which we describe next. 

\subsection{Learning the CNN word-embedding regressor}
\label{sec:sgfinetuning}

\noindent \textit{Step-1:} We fine-tune the VGG-16 object classification net~\cite{simonyan2014very} using Sketchy~\cite{sangkloy2016sketchy}, a large-scale sketch object dataset, for $125$-way classification corresponding to the $125$ categories present in the dataset. Let us denote the resulting fine-tuned net by $M_1$. 

\noindent \textit{Step-2:} $M_1$'s weights are used to initialize a VGG-16 net which is then fine-tuned for regressing word-embeddings corresponding to the $125$ category names of the Sketchy dataset. Specifically, we use the $500$-dimensional word-embeddings provided by the \texttt{word2vec} model trained on 1-billion Google News words~\cite{mikolov2013efficient}. Our choice is motivated by the open-ended nature of guess-words in Sketch-QA and the consequent need to capture semantic similarity between ground-truth and guess-words rather than perform exact matching. For the loss function w.r.t predicted word embedding $p$ and ground-truth embedding $g$, we consider [a] Mean Squared Loss : $\left\lVert p - g \right\rVert^2$ [b] Cosine Loss~\cite{qin2008query} : 1- $cos(p,g) = 1 - (p^T g/{\left\lVert p \right\rVert \left\lVert g \right\rVert})$ [c] Hinge-rank Loss~\cite{frome2013devise} : $max\lbrack 0, margin - {\hat{p}}^T\hat{g} + {\hat{p}}^T{\hat{h}}\rbrack$  where ${\hat{p}},{\hat{g}}$ are length-normalized versions of $p,g$ respectively and ${\hat{h}} (\neq {\hat{g}}$) corresponds to the normalized version of a randomly chosen category's word-embedding. The value for $margin$ is set to $0.1$ [d] Convex combination of Cosine Loss ($\text{CLoss}$) and Hinge-rank Loss ($\text{HLoss}$) : $\text{CLoss} + \lambda \text{HLoss}$. The predicted embedding $p$ is deemed a `correct' match if the set of its $k$-nearest word-embedding neighbors contains $g$. Overall, we found the convex combination loss with $\lambda=1$ (determined via grid search) to provide the best performance. Let us denote the resulting CNN regressor as $M_2$.

\noindent \textit{Step-3:} $M_2$ is now fine-tuned with randomly ordered sketches from training data sequences and corresponding word-embeddings. By repeating the grid search for the convex combination loss, we found $\lambda=1$ to once again provide the best performance  on the validation set. Note that in this case, ${\hat{h}}$ for Hinge-rank Loss corresponds to a word-embedding randomly selected from the entire word-embedding dictionary. Let us denote the fine-tuned CNN regressor by $M_3$.

As mentioned earlier, we use the $4096$-dimensional output from fc\textsubscript{7} layer of $M_3$ as the representation for each accumulated stroke image of sketch sequences.

\subsection{RNN training and evaluation}
\label{sec:sgfinetuning2}

\begin{figure*}[!t]
\centering
\includegraphics[width=\textwidth]{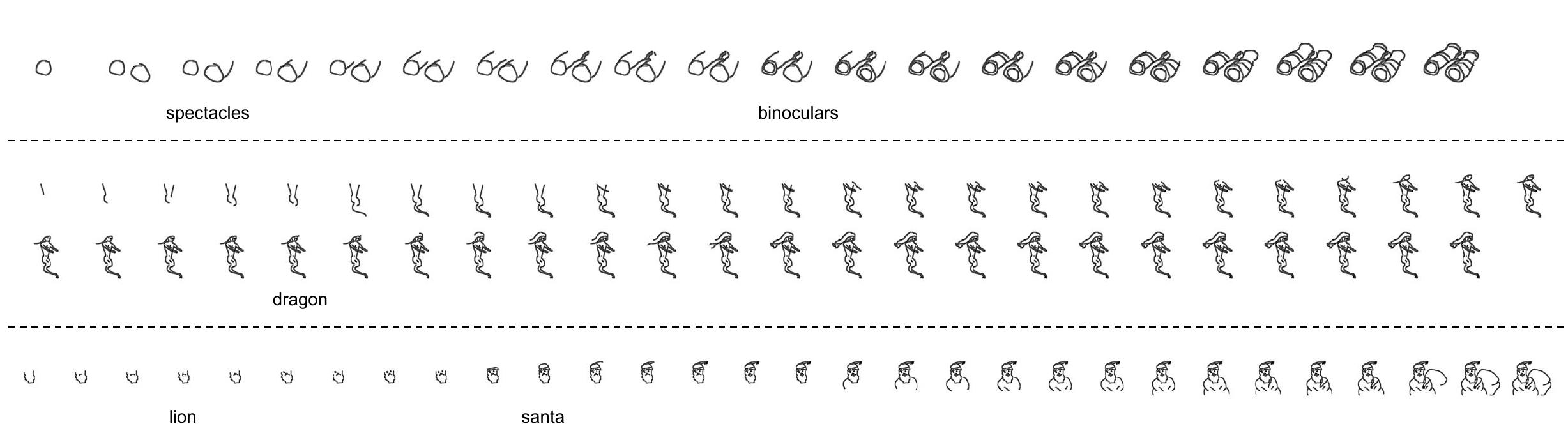}
\caption{Examples of guesses generated by our model on test set sequences.}
\label{fig:unifsampleguesses}
\end{figure*}

\noindent \textbf{RNN Training:} As with the CNN regressor, we configure the RNN to predict word-embeddings. For preliminary evaluation, we use only the portion of training sequences corresponding to guess-words. For each time-step, we use the same loss (convex combination of Cosine Loss and Hinge-rank Loss) determined to be best for the CNN regressor. We use LSTM~\cite{hochreiter1997long} as the specific RNN variant. For all the experiments, we use Adagrad optimizer~\cite{duchi2011adaptive} (with a starting learning rate of $0.01$) and early-stopping as the criterion for terminating optimization.

\noindent \textbf{Evaluation:} We use the $k$-nearest neighbor criteria mentioned above and examine performance for $k=1,2,3$. To determine the best configuration, we compute the proportion of `correct matches' on the subsequence of validation sequences containing guess-words. As a baseline, we also compute the sequence-level scores for the CNN regressor $M_3$. We average these per-sequence scores across the validation sequences. The results show that the CNN regressor performs reasonably well in spite of the overall complexity involved in regressing guess-word embeddings (see first row of Table \ref{tab:phase2}). However, this performance is noticeably surpassed by LSTM net, demonstrating the need to capture temporal context in modelling guess-word transitions. 

\begin{table}[!tbp]
\centering
\begin{tabularx}{\linewidth}{*1{Y|} *1{Y} *1{Y} *1{Y}}
\toprule
 LSTM & \multicolumn{3}{c}{Avg. sequence-level accuracy}\\
\cmidrule(l r){2-4}   
      & $1$ & $3$ & $5$ \\
\midrule
  --           & $52.77$ & $63.02$ & $66.40$      \\
$128$         & $54.13$ & $63.11$ & $66.25$                    \\
$256$         & $55.03$ & $63.79$ & $66.40$ \\
$512$         & $\bm{55.35}$ & $\bm{64.03}$ & $\bm{66.81}$                    \\
\bottomrule
\end{tabularx}
\caption{Sequence-level accuracies over the validation set are shown. In each sequence, only the portion with guess-words is considered for evaluation. The first row corresponds to $M_3$ CNN regressor. The first column shows the number of hidden units in the LSTM. The sequence level accuracies with k-nearest criteria applied to per-timestep guess predictions are shown for $k=1,3,5$.}
\label{tab:phase2} 
\end{table}

\section{Overall Results}
\label{sec:sgDeepGuesser-eval}

For the final model, we merge validation and training sets and re-train with the best architectural settings as determined by validation set performance (i.e. $M_3$ as the feature extraction CNN, LSTM with $512$ hidden units as the RNN component and convex combination of Cosine Loss and Hinge-rank Loss as the optimization objective). We report performance on the test sequences. 

\begin{figure}[!t]
\centering
\includegraphics[width=\linewidth,keepaspectratio]{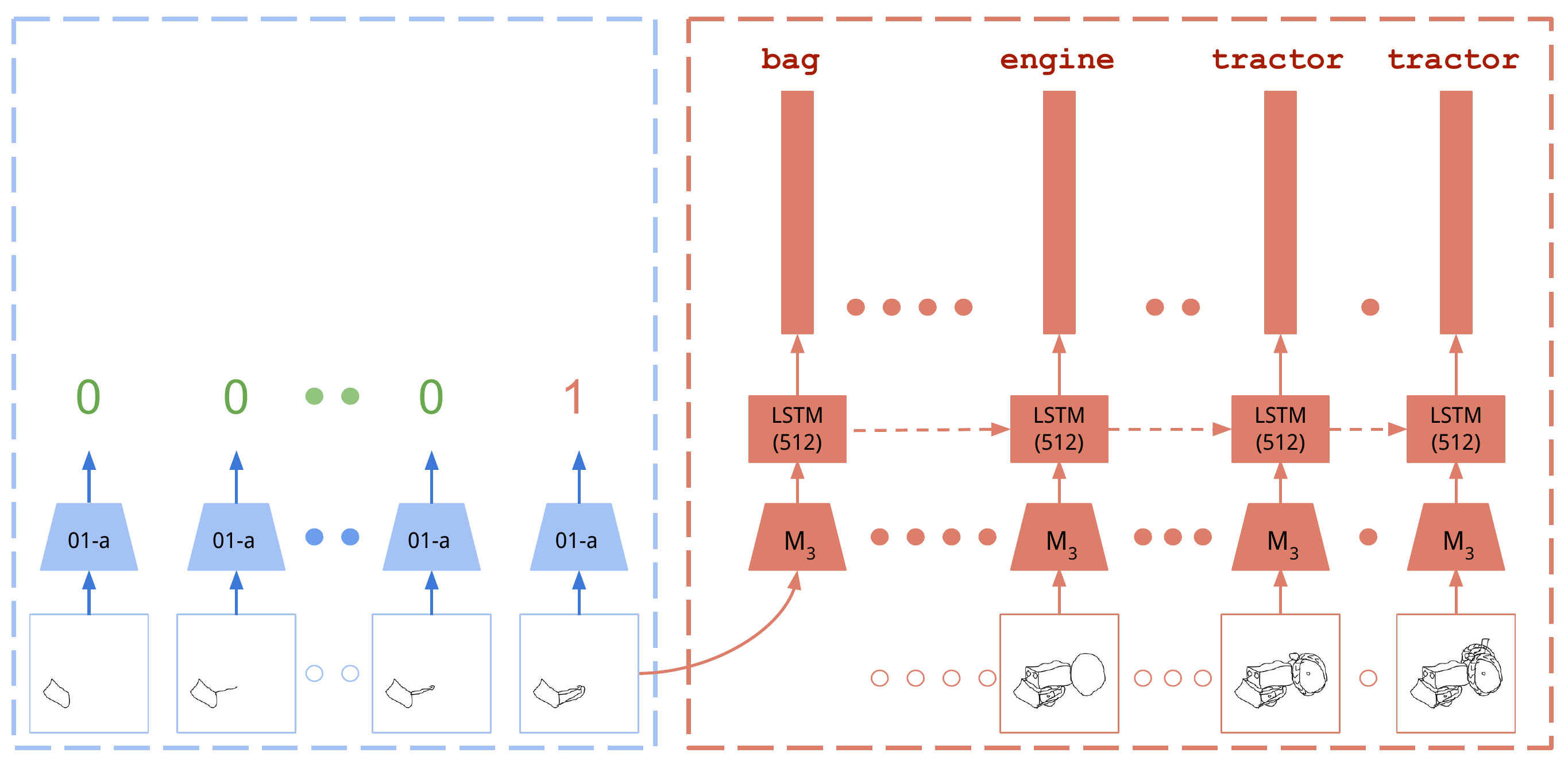}
\caption{Architecture for the two-phase baseline. The first phase (blue dotted line) is used to predict location of the transition to the word-guessing phase (output $1$). Starting from transition location, the second-phase (red dotted line) sequentially outputs word-embedding predictions until the end of stroke sequence.}
\label{fig:TwoPhase}
\end{figure}

The full-sequence scenario is considerably challenging since our model has the additional challenge of having to accurately determine when the word-guessing phase should begin. For this reason, we also design a two-phase architecture as an alternate baseline. In this baseline, the first phase predicts the most likely sequential location for `no guess'-to-first-guess transition. Conditioned on this location, the second phase predicts guess-word representations for rest of the sequence (see Figure \ref{fig:TwoPhase}). To retain focus, we only report performance numbers for the two-phase baseline. For a complete description of baseline architecture and related ablative experiments, please refer to Appendix \ref{sec:app:twoph}.

As can be observed in Table \ref{tab:overall}, our proposed word-guess model outperforms other baselines, including the two-phase baseline, by a significant margin. The reduction in long-range temporal contextual information, caused by splitting the original sequence into two disjoint sub-sequences, is possibly a reason for lower performance for the two-phase baseline. Additionally, the need to integrate sequential information is once again highlighted by the inferior performance of CNN-only baseline. We also wish to point out that $17\%$ of guesses in the test set are out-of-vocabulary words, i.e. guesses not present in train or validation set. Inspite of this, our model achieves high sequence-level accuracy, thus making the case for open-ended word-guessing models. 

Examples of guesses generated by our model on test set sketch sequences can be viewed in Figure \ref{fig:unifsampleguesses}.  

\noindent \textbf{Visual Turing Test:} As a subjective assessment of our model, we also conduct a Visual Turing Test. We randomly sample $K=200$ sequences from our test-set. For each of the model predictions, we use the nearest word-embedding as the corresponding guess. We construct two kinds of paired sequences $(s_i,h_i)$ and $(s_i,m_i)$ where $s_i$ corresponds to the $i$-th sketch stroke sequence ($1 \leqslant i \leqslant K$) and $h_i,m_i$ correspond to human and model generated guess sequences respectively. We randomly display the stroke-and-guess-word paired sequences to $20$ human judges with $10$ judges for each of the two sequence types. Without revealing the origin of guesses (human or machine), each judge is prompted ``Who produced these guesses?". 

\begin{table}[!tbp]
\centering
\begin{tabularx}{\linewidth}{*1{Y|} *1{Y} *1{Y} *1{Y}}
\toprule
 Architecture & \multicolumn{3}{c}{Avg. sequence-level accuracy}\\
\cmidrule(l r){2-4}   
      & $1$ & $3$ & $5$ \\
\midrule
$M_3$ (CNN)   & $43.61$ & $51.54$ & $54.18$      \\
Two-phase     & $46.33$ & $52.08$ & $54.46$   \\ 
Proposed      & $\bm{62.04}$ & $\bm{69.35}$ & $\bm{71.11}$      \\
\bottomrule
\end{tabularx}
\caption{Overall average sequence-level accuracy on test set are shown for guessing models (CNNs only baseline [first row], two-phase baseline [second] and our proposed model [third]).}
\label{tab:overall} 
\end{table}

The judges entered their ratings on a $5$-point Likert scale (`Very likely a machine', `Either is equally likely','Very likely a human'). To minimize selection bias, the scale ordering is reversed for half the subjects~\cite{chan1991response}. For each sequence $i, 1 \leqslant i \leqslant K$, we first compute the mode (${\mu}^\mathcal{H}_i$ (human guesses), ${\mu}^\mathcal{M}_i$ (model guesses)) of the $10$ ratings by guesser type. To determine the statistical significance of the ratings, we additionally analyze the $K$ rating pairs ($({\mu}^\mathcal{H}_i,{\mu}^\mathcal{M}_i), 1 \leqslant i \leqslant K$) using the non-parametric Wilcoxon Signed-Rank test~\cite{10.2307/3001968}. 

\begin{figure}[!t]
\centering
\includegraphics[width=\linewidth]{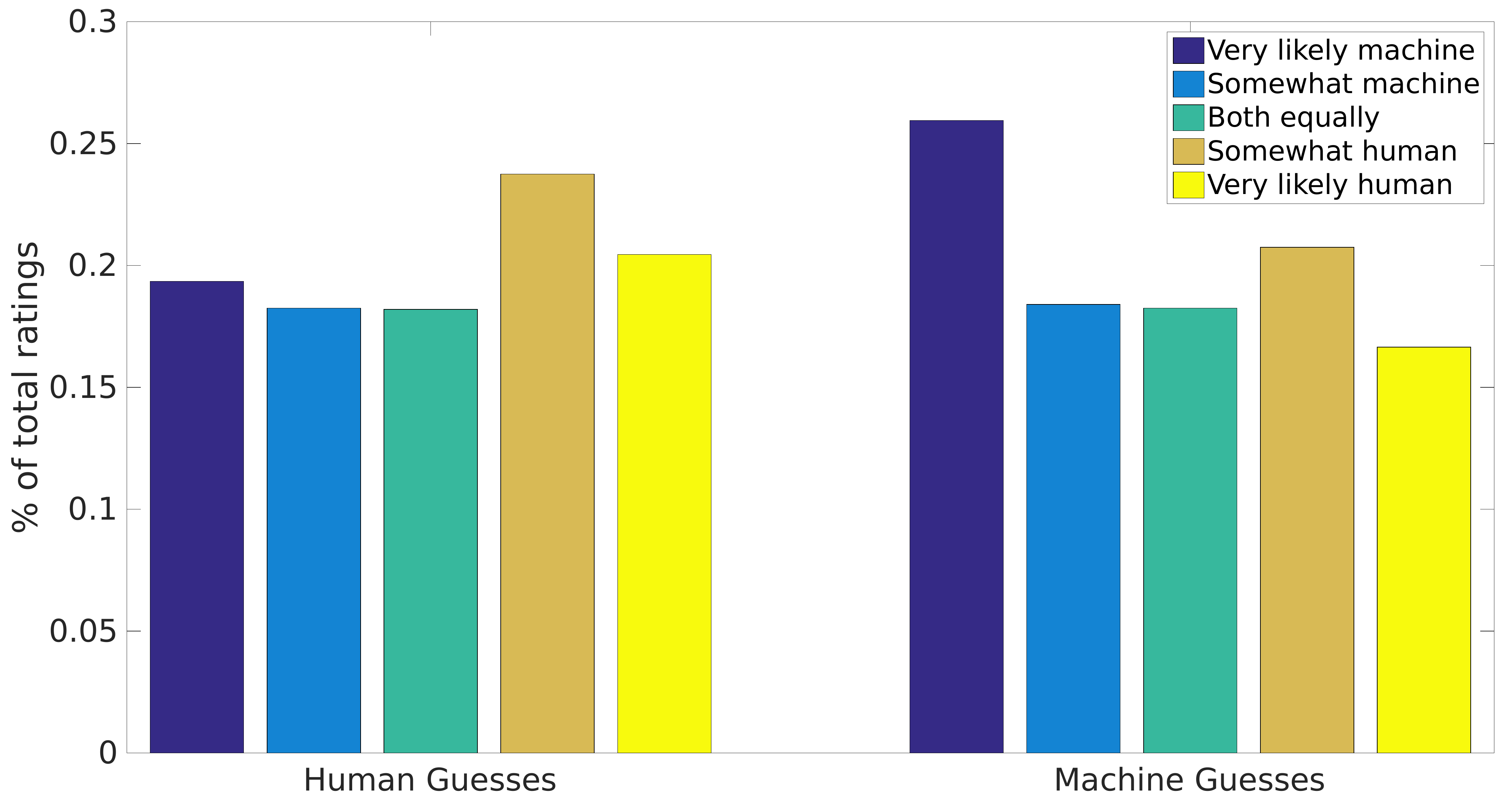}
\caption{Distribution of ratings for human and machine-generated guesses.}
\label{fig:guessdist}
\end{figure}

When we study the distribution of ratings (Figure \ref{fig:guessdist}), the human subject-based guesses from \textsc{WordGuess-160} seem to be clearly identified as such -- the two most frequent rating levels correspond to `human'. The non-trivial frequency of `machine' ratings reflects the ambiguity induced not only by sketches and associated guesses, but also by the possibility of machine being an equally viable generator. For the model-generated guesses, many could be identified as such, indicating the need for more sophisticated guessing models. This is also evident from the Wilcoxon Signed-Rank test which indicates a significant effect due to the guesser type ($p=0.005682,Z=2.765593$). Interestingly, the second-most preferred rating for model guesses is `human', indicating a degree of success for the proposed model. 

\section{Related Work}
\label{sec:sgrelatedwork}

Beyond its obvious entertainment value, Pictionary involves a number of social~\cite{wortham2006adapting,mayra2007}, collaborative~\cite{fay2013bootstrap,Groen20121575} and cognitive~\cite{dake1995visual,kievit2011semantic} aspects which have been studied by researchers. In an attempt to find neural correlates of creativity, Saggar et al.~\cite{saggar2015pictionary} analyze fMRI data of participants instructed to draw sketches of Pictionary `action' words (E.g. ``Salute",``Snore"). In our approach, we ask subjects to guess the word instead of drawing the sketch for a given word. Also, our sketches correspond to nouns (objects). 

Human-elicited text-based responses to visual content, particularly in game-like settings, have been explored for object categorization~\cite{von2004labeling,branson2010visual}. However, the visual content is static and does not accumulate sequentially, unlike our case. The work of Ullman et al.~\cite{ullman2016atoms} on determining minimally recognizable image configurations also bears mention. Our approach is complementary to theirs in the sense that we incrementally add stroke content (bottom-up) while they incrementally reduce image content (top-down).  

In recent times, deep architectures for sketch recognition~\cite{yu2015sketch,seddati2015deepsketch,Sarvadevabhatla:2016:EMR:2964284.2967220} have found great success. However, these models are trained to output a single, fixed label regardless of the intra-category variation. In contrast, our model, trained on actual human guesses, naturally exhibits human-like variety in its responses (e.g. a sketch can be guessed as `aeroplane' or `warplane' based on evolution of stroke-based appearance). Also, our model solves a much more complex temporally-conditioned, multiple word-embedding regression problem. Another important distinction is that our dataset (\textsc{WordGuess-160}) contains incorrect guesses which usually arise due to ambiguity in sketched depictions. Such `errors' are normally considered undesirable, but we deliberately include them in the training phase to enable realistic mimicking. This in turn requires our model to implicitly capture the subtle, fine-grained variations in sketch quality -- a situation not faced by existing approaches which simply optimize for classification accuracy.

Our dataset collection procedure is similar to the one employed by Johnson et al.~\cite{johnson2009games} as part of their Pictionary-style game Stellasketch. However, we do not let the subject choose the object category. Also, our subjects only provide guesses for stroke sequences of existing sketches and not for sketches being created in real-time. Unfortunately, the Stellasketch dataset is not available publicly for further study.

It is also pertinent to compare our task and dataset with QuickDraw, a large-scale sketch collection initiative by Google (\url{https://github.com/googlecreativelab/quickdraw-dataset}). The QuickDraw task generates a dataset of object sketches. In contrast, our task SketchQA results in a dataset of human-generated guess words. In QuickDraw, a sketch is associated with a single, fixed category. In SketchQA, a sketch from an existing dataset is explicitly associated with a list of multiple guess words. In SketchQA, the freedom provided to human guessers enables sketches to have arbitrarily fine-grained labels (e.g. `airplane', `warplane',`biplane'). However, QuickDraw's label set is fixed. Finally, our dataset (\textsc{WordGuess-160}) captures a rich sequence of guesses in response to accumulation of sketch strokes. Therefore, it can be used to train human-like guessing models. QuickDraw's dataset, lacking human guesses, is not suited for this purpose. 

Our computational model employs the Long Short Term Memory (LSTM)~\cite{hochreiter1997long} variant of Recurrent Neural Networks (RNNs). LSTM-based frameworks have been utilized for tasks involving temporally evolving content such as as video captioning~\cite{donahue2015long,venugopalan2015sequence} and action recognition ~\cite{yeung2015every,7299101,ma2016learning}. Our model not only needs to produce human-like guesses in response to temporally accumulated content, but also has the additional challenge of determining how long to `wait' before initiating the guessing process. Once the guessing phase begins, our model typically outputs multiple answers. These per-time-step answers may even be unrelated to each other. This paradigm is different from a setup wherein a single answer constitutes the output. Also, the output of RNN in aforementioned approaches is a soft-max distribution over \textit{all} the words from a fixed dictionary. In contrast, we use a regression formulation wherein the RNN outputs a word-embedding prediction at each time-step. This ensures scalability with increase in vocabulary and better generalization since our model outputs predictions in a constant-dimension vector space. ~\cite{lev2016rnn} adopt a similar regression formulation to obtain improved performance for image annotation and action recognition. 

Since our model aims to mimic human-like guessing behavior, a subjective evaluation of generated guesses falls within the ambit of a Visual Turing Test~\cite{geman2015visual,malinowski2014towards,gao2015you}. However, the free-form nature of guess-words and the ambiguity arising from partial stroke information make our task uniquely more challenging.

\section{Discussion and Conclusion}

We have introduced a novel guessing task called Sketch-QA to crowd-source Pictionary-style open-ended guesses for object line sketches as they are drawn. The resulting dataset, dubbed \textsc{GuessWord-160}, contains $16624$ guess sequences of $1108$ subjects across $160$ object categories. We have also introduced a novel computational model which produces open-ended guesses and analyzed its performance on \textsc{GuessWord-160} dataset for challenging on-line Pictionary-style guessing tasks. 

In addition to the computational model, our dataset \textsc{GuessWord-160} can serve researchers studying human perceptions of iconic object depictions. Since the guess-words are paired with object depictions, our data can also aid graphic designers and civic planners in creation of meaningful logos and public signage. This is especially important since incorrectly perceived depictions often result in inconvenience, mild amusement, or in extreme cases, end up deemed offensive. Yet another potential application domain is clinical healthcare. \textsc{GuessWord-160} consists of partially drawn objects and corresponding guesses across a large number of categories. Such data could be useful for neuro psychiatrists to characterize conditions such as visual agnosia: a disorder in which subjects exhibit impaired object recognition capabilities~\cite{agnosia}.

In future, we wish to also explore computational models for optimal guessing, i.e. models which aim to guess the sketch category as early and as correctly as possible.  In the futuristic context mentioned at the beginning (Figure \ref{fig:robotpictionary}), such models would help the robot contribute as a productive team-player by correctly guessing its team-member's sketch as early as possible. In our dataset, each stroke sequence was shown only to a single subject and therefore, is associated with a single corresponding sequence of guesses. This shortcoming is to be mitigated in future editions of Sketch-QA. A promising approach for data collection would be to use digital whiteboards, high-quality microphones and state-of-the-art speech recognition software to collect realistic paired stroke-and-guess data from Pictionary games in home-like settings~\cite{sigurdsson2016hollywood}. It would also be worthwhile to consider Sketch-QA beyond object names (`nouns') and include additional lexical types (e.g. action-words and abstract phrases). We believe the resulting data, coupled with improved versions of our computational models, could make the scenario from Figure \ref{fig:robotpictionary} a reality one day.

\appendices

\section{Two-Phase Baseline Model}
\label{sec:app:twoph}

In this section, we present the architectural design and related evaluation experiments of the two-phase baseline originally mentioned  in Section \ref{sec:sgDeepGuesser-eval}.

Typically, a guess sequence contains two distinct phases. In the first phase, no guesses are provided by the subject since the accumulated strokes provide insufficient evidence. At a later stage, the subject feels confident enough to provide the first guess. Thus, the location of this first guess (within the overall sequence) is the starting point for the second phase. The first phase (i.e. no guesses) offers no usable guess-words. Therefore, rather than tackling both the phases within a single model, we adopt a divide-and-conquer approach. We design this baseline to first predict the phase transition location (i.e. where the first guess occurs). Conditioned on this location, the model predicts guess-word representations for rest of the sequence (see Figure \ref{fig:TwoPhase}).

In the two-phase model and the model described in the main paper, the guess-word generator is a common component. The guess-word generation model is already described in the main paper (Section \ref{sec:sgcompmodels}). For the remainder of the section, we focus on the first phase of the two-phase baseline.

Consider a typical guess sequence $\mathbb{G_I}$. Suppose the first phase (`no guesses') corresponds to an initial sub-sequence of length $k$. The second phase then corresponds to the remainder sub-sequence of length $(N-k)$. Denoting `no guess' as $0$ and a guess-word as $1$, $\mathbb{G_I}$ is transformed to a binary sequence $\mathbb{B_I} = \lbrack (0,0 \ldots k \text{ times }) (1,1 \ldots (N-k) \text{ times}) \rbrack$. Therefore, the objective for the Phase I model is to correctly predict the transition index i.e. $(k+1)$. 

\subsection{Phase I model (Transition prediction)}
\label{sec:appphase1}

\begin{table*}[!h]
\begin{tabularx}{\linewidth}{*3{Y|} *2{Y} *1{Y}}
\toprule
CNN model & LSTM & Loss& \multicolumn{3}{|c}{Window width}\\
\cmidrule(l r){4-6}   
& & & $1$ & $3$ & $5$ \\
\midrule
01        & --         & CCE      & $17.37$ & $36.57$ & $49.67$ \\
01-a      & --         & CCE      & $\bm{20.45}$ & $41.22$ & $\bm{54.91}$ \\
01-a      & 64         & Seq      & $17.30$ & $38.40$ & $52.75$ \\
01-a      & 128        & Seq      & $18.94$ & $39.25$ & $53.47$ \\
01-a      & 256        & Seq      & $18.68$ & $40.04$ & $53.41$ \\
01-a      & 512        & Seq      & $18.22$ & $39.45$ & $54.78$ \\
01-a      & 128        & wSeq    & $19.20$ & $\bm{41.48}$ & $\bm{55.64}$ \\
01-a      & 128        & mRnk    & $18.87$ & $37.88$ & $52.23$ \\
\bottomrule
\end{tabularx}
\caption{The transition location prediction accuracies for various Phase I architectures are shown. 01 refers to the binary output CNN model pre-trained for feature extraction. 01-a refers to the 01 CNN model with $160$-way auxiliary classification. The last two rows correspond to test set accuracies of the best CNN and LSTM configurations. For the `Loss' column, CCE = Categorical-cross entropy, Seq = Average sequence loss, wSeq = Weighted sequence loss, mRnk = modified Ranking Loss. The results are shown for `Window width' sized windows centered on ground-truth transition location. The rows below dotted line show performance of best CNN and LSTM models on test sequences.}
\label{tab:appphase1}
\end{table*}

Two possibilities exist for Phase-I model. The first possibility is to train a CNN model using sequence members from $\mathbb{I},\mathbb{B_I}$ pairs for binary (Guess/No Guess) classification and during inference, repeatedly apply the CNN model on successive time-steps, stopping when the CNN model outputs $1$ (indicating the beginning of guessing phase). The second possibility is to train an RNN and during inference, stop unrolling when a $1$ is encountered. We describe the setup for CNN model first.

\subsubsection{CNN model}
For the CNN model, we fine-tune VGG-16 object classification model~\cite{simonyan2014very} using Sketchy~\cite{sangkloy2016sketchy} as in the proposed model. The fine-tuned model is used to initialize another VGG-16 model, but with a $256$-dimensional bottleneck layer introduced after fc\textsubscript{7} layer. Let us denote this model as $Q_1$. 

\subsubsection{Sketch representation}

 As feature representations, we consider two possibilities:app [a] $Q_1$ is fine-tuned for $2$-way classification (Guess/No Guess). The $256$-dimensional output from final fully-connected layer forms the feature representation. [b] The architecture in option [a] is modified by having $160$-way class prediction as an additional, auxiliary task. This choice is motivated by the possibility of encoding category-specific transition location statistics within the $256$-dimensional feature representation (see Figure \ref{fig:firstguessloc}). The two losses corresponding to the two outputs ($2$-way and $160$-way classification) of the modified architecture are weighted equally during training. 

\noindent \textbf{Loss weighting for imbalanced label distributions:} When training the feature extraction CNN ($Q_1$) in Phase-I, we encounter imbalance in the distribution of no-guesses ($0$s) and guesses ($1$s). To mitigate this, we employ class-based loss weighting~\cite{eigen2015predicting} for the binary classification task. Suppose the number of no-guess samples is $n$ and the number of guess samples is $g$. Let $\mu = \frac{n+g}{2}$. The weights for the classes are computed as $w_0 = \frac{\mu}{f_0}$ where $f_0 = \frac{n}{(n+g)}$ and $w_1 = \frac{\mu}{f_1}$ where $f_1 = \frac{g}{(n+g)}$. The binary cross-entropy loss is then computed as: 

\begin{equation}
\resizebox{0.5\textwidth}{!}{$
\mathcal{L}(P,G) = \displaystyle \sum_{x \in \mathcal{X}_{train}} - w_x \lbrack g_x log (p_x) + (1-g_x) log (1 - p_x) \rbrack$
}
\end{equation}

where $g_x,p_x$ stand for ground-truth and prediction respectively and $w_x=w_0$ when $x$ is a no-guess sample and $w_x=w_1$ otherwise. For our data, $w_0=1.475$ and $w_1=0.765$, thus appropriately accounting for the relatively smaller number of no-guess samples in our training data.

A similar procedure is also used for weighting losses when the $160$-way auxiliary classifier variant of $Q_1$ is trained. In this case, the weights are determined by the per-object category distribution of the training sequences. Experimentally, $Q_1$ with auxiliary task shows better performance -- see first two rows of Table \ref{tab:appphase1}. 

\subsubsection{LSTM setup}

 We use the $256$-dimensional output of the $Q_1$-auxiliary CNN as the per-timestep sketch representation fed to the LSTM model. To capture the temporal evolution of the binary sequences, we configure the LSTM to output a binary label $B_t \in \{0,1\}$ for each timestep $t$. For the LSTM, we explored variations in number of hidden units ($64,128,256,512$). The weight matrices are initialized as orthogonal matrices with a gain  factor of $1.1$~\cite{DBLP:journals/corr/SaxeMG13} and the forget gate bias is set to $1$. For training the LSTMs, we use the average sequence loss, computed as the average of the per-time-step binary cross-entropy losses. The loss is regularized by a standard $L_2$-weight norm weight-decay parameter ($\alpha=0.0005$). For optimization, we use Adagrad with a learning rate of $5 \times 10^{-5}$ and the momentum term set to $0.9$. The gradients are clipped to $5.0$ during training. For all LSTM experiments, we use a mini-batch size of $1$.

\subsubsection{LSTM Loss function variants} 

The default sequence loss formulation treats all time-steps of the sequence equally. Since we are interested in accurate localization of transition point, we explored the following modifications of the default loss for LSTM:

\noindent \textit{Transition weighted loss:} To encourage correct prediction at the transition location, we explored a weighted version of the default sequence-level loss. Beginning at the transition location, the per-timestep losses on either side of the transition are weighted by an exponentially decaying factor $e^{-\alpha ( 1 - [t/(k+1)]^{s})}$ where $s = 1$ for time-steps $\lbrack 1,k \rbrack$,  $s=-1$ for $\lbrack k+2,N \rbrack$. Essentially, the loss at the transition location is weighted the most while the losses for other locations are downscaled by weights less than $1$ -- the larger the distance from transition location, the smaller the weight. We tried various values for $\alpha$. The localization accuracy can be viewed in Table \ref{tab:appwtloss}. Note that the weighted loss is added to the original sequence loss during actual training. 

\begin{table}[!tbp]
\centering
\begin{tabularx}{\linewidth}{*3{Y|} *2{Y} *1{Y}}
\toprule
CNN model & LSTM & $\alpha$ & \multicolumn{3}{|c}{Window width}\\
\cmidrule(l r){4-6}   
& & & $1$ & $3$ & $5$ \\
\midrule
01-a      & 128        &  5       & $19.00$      & $41.55$      & $55.44$ \\
01-a      & 128        & 7        & $19.20$      & $\bm{41.48}$ & $\bm{54.85}$\\
01-a      & 128        &  10      & $18.48$      & $40.10$      & $54.06$ \\
\bottomrule
\end{tabularx}
\caption{\label{tab:appwtloss} Weighted loss performance for various values of $\alpha$.}
\end{table}

\noindent \textit{Modified ranking loss:} We want the model to prevent occurrence of premature or multiple transitions. To incorporate this notion, we use the ranking loss formulation proposed by Ma et al.~\cite{ma2016learning}. Let us denote the loss at time step $t$ as $\mathcal{L}^t_c$ and the softmax score for the ground truth label $y_t$ as $p_t^{y_t}$. We shall refer to this as detection score. In our case, for the Phase-I model, $\mathcal{L}^t_c$ corresponds to the binary cross-entropy loss. The overall loss at time step $t$ is modified as: 

\begin{align}
\mathcal{L}^t = \lambda_s \mathcal{L}^t_c + \lambda_r \mathcal{L}^t_r
\end{align}

We want the Phase-I model to produce monotonically non-decreasing softmax values for no-guesses and guesses as it progresses more into the sub-sequence. In other words, if there is no transition at time $t$, i.e. $y_{t} = y_{t-1}$, then we want the current detection score to be no less than any previous detection score. Therefore, for this situation, the ranking loss is computed as:

\begin{align}
    \mathcal{L}_r^t = max(0,p_t^{*y_t} - p_t^{y_t})
\end{align}

\noindent where 

\begin{align}
    p_t^{*y_t}    = \max_{t^{'} \in \lbrack t_{s},t-1 \rbrack} p_{t^{'}}^{y_{t}}    
\end{align}

\noindent where $t_s$ corresponds to time step $1$ when $y_t=0$ (No Guesses) or $t_s = t_p$ (starting location of Guessing).

\noindent If time-step $t$ corresponds to a transition, i.e. $y_{t} \neq y_{t-1}$, we want the detection score of previous phase (`No Guess') to be as small as possible (ideally $0$). Therefore, we compute the ranking loss as:

\begin{align}
    \mathcal{L}_r^t = p_t^{y_{t-1}}
\end{align}

\noindent During training, we use a convex combination of sequence loss and the ranking loss with the loss weighting determined by grid search over $\lambda_{s-r}$ (see Table \ref{tab:apprankloss}). From our experiments, we found the transition weighted loss to provide the best performance (Table \ref{tab:appphase1}). 

\begin{table*}[!h]
\centering
\begin{tabularx}{\linewidth}{*3{Y|} *2{Y} *1{Y}}
\toprule
CNN model & LSTM & $\lambda_s,\lambda_r$ & \multicolumn{3}{|c}{Window width}\\
\cmidrule(l r){4-6}   
& & & $1$ & $3$ & $5$ \\
\midrule
01-a      & 128        &  $0.5,1.0$       & $17.43$      & $35.26$      & $48.23$ \\
01-a      & 128        & $1,1$        & $18.41$      & $39.45$ & $53.08$\\
01-a      & 128        &  $1,0.5$      & $18.87$      & $37.88$      & $52.23$ \\
\bottomrule
\end{tabularx}
\caption{\label{tab:apprankloss} Ranking loss performance for various weighting of sequence loss and rank loss.}
\end{table*}

\begin{table*}[!t]
\small
\centering
\resizebox{\linewidth}{!}{
\begin{tabularx}{\linewidth}{*2{Y|} *1{Z} *1{V|} *1{Z} *1{V|} *1{Z} *1{V}}
%\begin{tabularx}{\linewidth}{*2{Y|} *1{Y} *1{Y|} *1{Y} *1{Y|} *1{Y} *1{Y}}
\toprule
 P-I  & P-II & \multicolumn{6}{c}{Average sequence-level accuracy} \\
  \cmidrule(l  r){3-8}   
     &      & \multicolumn{2}{c|}{$k=1$} & \multicolumn{2}{c|}{$k=3$} & \multicolumn{2}{c}{$k=5$} \\
  \cmidrule(l r){3-8}   
      &     & P-II only   &  Full   &  P-II only   & Full  & P-II only  & Full   \\
\midrule
 01-a                                   & $M_3$  & $54.06$ & $43.61$ & $64.11$ & $51.54$ & $66.85$ &  $54.18$ \\ 
 Unified      & Unified  & $46.35$ & $\bm{62.04}$ & $56.45$ & $\bm{69.35}$ & $59.30$ & $\bm{71.11}$ \\ 
 01-a           & R2\textsubscript{5}   & $\bm{57.05}$ & $46.33$ & $\bm{64.76}$ & $52.08$ & $\bm{67.19}$ & $54.46$ \\  
\bottomrule
\end{tabularx} 
}
\caption{\label{tab:appoverall} Overall average sequence-level accuracy on test set are shown for guessing models (CNNs only baseline [first row], Unified [second], Two Phased [third]). R2\textsubscript{5} corresponds to best Phase-II LSTM model.}
\end{table*}

\subsubsection{Evaluation}

At inference time, the accumulated stroke sequence is processed sequentially by Phase-I model until it outputs a $1$ which marks the beginning of Phase-II. Suppose the predicted transition index is $p$ and ground-truth index is $g$. The prediction is deemed correct if $p \in [g-\delta,g+\delta]$ where $\delta$ denotes half-width of a window centered on $p$. For our experiments, we used $\delta \in \{0,1,2\}$. The results (Table \ref{tab:appphase1}) indicate that the $Q_1$-auxiliary CNN model outperforms the best LSTM model by a very small margin. The addition of weighted sequence loss to the default version plays a crucial role in the latter (LSTM model) since the default version does not explicitly optimize for the transition location. Overall, the large variation in sequence lengths and transition locations explains the low performance for exact ($k=1$) localization. Note, however, that the performance improves considerably when just one to two nearby locations are considered for evaluation ($k=3,5$).

During inference, the location predicted by Phase-I model is used as the starting point for Phase-II (word guessing). We do not describe Phase-II model since it is virtually identical in design as the model described in the main paper (Section \ref{sec:sgcompmodels}).

\subsection{Overall Results}
\label{sec:appDeepGuesser-eval}

To determine overall performance, we utilize the best architectural settings as determined by validation set performance. We then merge validation and training sets, re-train the best models and report their performance on the test set. As the overall performance measure, we report two items on the test set -- [a] \textit{P-II}: the fraction of correct matches with respect to the subsequence corresponding to ground-truth word guesses. In other words, we assume $100\%$ accurate localization during Phase I and perform Phase II inference beginning from the ground-truth location of the first guess. [b] \textit{Full:} We use Phase-I model to determine transition location. Note that depending on predicted location, it is possible that we obtain word-embedding predictions when the ground-truth at the corresponding time-step corresponds to `no guess'. Regarding such predictions as mismatches, we compute the fraction of correct matches for the full sequence. As a baseline model (first row of Table \ref{tab:appoverall}), we use outputs of the best performing per-frame CNNs from Phase I and Phase II. 

The results (Table \ref{tab:appoverall}) show that the Unified model outperforms Two-Phased model by a significant margin. For Phase-II model, the objective for CNN (whose features are used as sketch representation) and LSTM are the same. This is not the case for Phase-I model. The reduction in long-range temporal contextual information, caused by splitting the original sequence into two disjoint sub-sequences, is possibly another reason for lower performance of the Two-Phased model. 


\bibliographystyle{IEEEtran}

\begin{thebibliography}{10}
\providecommand{\url}[1]{#1}
\csname url@samestyle\endcsname
\providecommand{\newblock}{\relax}
\providecommand{\bibinfo}[2]{#2}
\providecommand{\BIBentrySTDinterwordspacing}{\spaceskip=0pt\relax}
\providecommand{\BIBentryALTinterwordstretchfactor}{4}
\providecommand{\BIBentryALTinterwordspacing}{\spaceskip=\fontdimen2\font plus
\BIBentryALTinterwordstretchfactor\fontdimen3\font minus
  \fontdimen4\font\relax}
\providecommand{\BIBforeignlanguage}[2]{{%
\expandafter\ifx\csname l@#1\endcsname\relax
\typeout{** WARNING: IEEEtran.bst: No hyphenation pattern has been}%
\typeout{** loaded for the language `#1'. Using the pattern for}%
\typeout{** the default language instead.}%
\else
\language=\csname l@#1\endcsname
\fi
#2}}
\providecommand{\BIBdecl}{\relax}
\BIBdecl

\bibitem{tesauro1994td}
G.~Tesauro, ``{TD}-gammon, a self-teaching backgammon program, achieves
  master-level play,'' \emph{Neural Computation}, vol.~6, no.~2, pp. 215--219,
  1994.

\bibitem{DBLP:conf/aaai/1997w6}
\emph{Deep Blue Versus Kasparov: The Significance for Artificial
  Intelligence}.\hskip 1em plus 0.5em minus 0.4em\relax {AAAI} Press, 1997.

\bibitem{silver2016mastering}
D.~Silver \emph{et~al.}, ``Mastering the game of {G}o with deep neural networks
  and tree search,'' \emph{Nature}, vol. 529, no. 7587, pp. 484--489, 2016.

\bibitem{chen2015mind}
X.~Chen and C.~Lawrence~Zitnick, ``Mind's eye: A recurrent visual
  representation for image caption generation,'' in \emph{CVPR}, 2015, pp.
  2422--2431.

\bibitem{venugopalan2015sequence}
S.~Venugopalan, M.~Rohrbach, J.~Donahue, R.~Mooney, T.~Darrell, and K.~Saenko,
  ``Sequence to sequence-video to text,'' in \emph{CVPR}, 2015, pp. 4534--4542.

\bibitem{xu2015show}
K.~Xu, J.~Ba, R.~Kiros, K.~Cho, A.~C. Courville, R.~Salakhutdinov, R.~S. Zemel,
  and Y.~Bengio, ``Show, attend and tell: Neural image caption generation with
  visual attention.'' in \emph{ICML}, vol.~14, 2015, pp. 77--81.

\bibitem{antol2015vqa}
S.~Antol, A.~Agrawal, J.~Lu, M.~Mitchell, D.~Batra, C.~Lawrence~Zitnick, and
  D.~Parikh, ``{VQA}: Visual question answering,'' in \emph{ICCV}, 2015, pp.
  2425--2433.

\bibitem{Xu2016}
H.~Xu and K.~Saenko, ``Ask, attend and answer: Exploring question-guided
  spatial attention for visual question answering,'' in \emph{ECCV}.\hskip 1em
  plus 0.5em minus 0.4em\relax Springer, 2016, pp. 451--466.

\bibitem{ren2015exploring}
M.~Ren, R.~Kiros, and R.~Zemel, ``Exploring models and data for image question
  answering,'' in \emph{NIPS}, 2015, pp. 2953--2961.

\bibitem{eitz2012humans}
M.~Eitz, J.~Hays, and M.~Alexa, ``How do humans sketch objects?'' \emph{ACM
  Trans. on Graphics}, vol.~31, no.~4, p.~44, 2012.

\bibitem{Schneider:2014:SCC:2661229.2661231}
R.~G. Schneider and T.~Tuytelaars, ``Sketch classification and
  classification-driven analysis using fisher vectors,'' \emph{ACM Trans.
  Graph.}, vol.~33, no.~6, pp. 174:1--174:9, Nov. 2014.

\bibitem{halacsy2007hunpos}
P.~Hal{\'a}csy, A.~Kornai, and C.~Oravecz, ``{H}un{P}os: an open source trigram
  tagger,'' in \emph{Proc. ACL on interactive poster and demonstration
  sessions}, 2007, pp. 209--212.

\bibitem{enchant}
D.~Lachowicz, ``Enchant spellchecker library,'' 2010.

\bibitem{miller1995wordnet}
G.~A. Miller, ``Wordnet: a lexical database for english,'' \emph{Communications
  of the ACM}, vol.~38, no.~11, pp. 39--41, 1995.

\bibitem{wu1994verbs}
Z.~Wu and M.~Palmer, ``Verbs semantics and lexical selection,'' in
  \emph{ACL}.\hskip 1em plus 0.5em minus 0.4em\relax Association for
  Computational Linguistics, 1994, pp. 133--138.

\bibitem{Sarvadevabhatla:2016:EMR:2964284.2967220}
R.~K. Sarvadevabhatla, J.~Kundu, and V.~B. Radhakrishnan, ``Enabling my robot
  to play pictionary: Recurrent neural networks for sketch recognition,'' in
  \emph{ACMMM}, 2016, pp. 247--251.

\bibitem{simonyan2014very}
K.~Simonyan and A.~Zisserman, ``Very deep convolutional networks for
  large-scale image recognition,'' \emph{arXiv preprint arXiv:1409.1556}, 2014.

\bibitem{sangkloy2016sketchy}
P.~Sangkloy, N.~Burnell, C.~Ham, and J.~Hays, ``The sketchy database: learning
  to retrieve badly drawn bunnies,'' \emph{ACM Transactions on Graphics (TOG)},
  vol.~35, no.~4, p. 119, 2016.

\bibitem{mikolov2013efficient}
T.~Mikolov, K.~Chen, G.~Corrado, and J.~Dean, ``Efficient estimation of word
  representations in vector space,'' \emph{arXiv preprint arXiv:1301.3781},
  2013.

\bibitem{qin2008query}
T.~Qin, X.-D. Zhang, M.-F. Tsai, D.-S. Wang, T.-Y. Liu, and H.~Li,
  ``Query-level loss functions for information retrieval,'' \emph{Information
  Processing \& Management}, vol.~44, no.~2, pp. 838--855, 2008.

\bibitem{frome2013devise}
A.~Frome, G.~S. Corrado, J.~Shlens, S.~Bengio, J.~Dean, T.~Mikolov
  \emph{et~al.}, ``Devise: A deep visual-semantic embedding model,'' in
  \emph{NIPS}, 2013, pp. 2121--2129.

\bibitem{hochreiter1997long}
S.~Hochreiter and J.~Schmidhuber, ``Long short-term memory,'' \emph{Neural
  Computation}, vol.~9, no.~8, pp. 1735--1780, 1997.

\bibitem{duchi2011adaptive}
J.~Duchi, E.~Hazan, and Y.~Singer, ``Adaptive subgradient methods for online
  learning and stochastic optimization,'' \emph{JMLR}, vol.~12, no. Jul, pp.
  2121--2159, 2011.

\bibitem{chan1991response}
J.~C. Chan, ``Response-order effects in likert-type scales,'' \emph{Educational
  and Psychological Measurement}, vol.~51, no.~3, pp. 531--540, 1991.

\bibitem{10.2307/3001968}
\BIBentryALTinterwordspacing
F.~Wilcoxon, ``Individual comparisons by ranking methods,'' \emph{Biometrics
  Bulletin}, vol.~1, no.~6, pp. 80--83, 1945. [Online]. Available:
  \url{http://www.jstor.org/stable/3001968}
\BIBentrySTDinterwordspacing

\bibitem{wortham2006adapting}
T.~B. Wortham, ``Adapting common popular games to a human factors/ergonomics
  course,'' in \emph{Proc. Human Factors and Ergonomics Soc. Annual Meeting},
  vol.~50.\hskip 1em plus 0.5em minus 0.4em\relax SAGE, 2006, pp. 2259--2263.

\bibitem{mayra2007}
F.~M{\"a}yr{\"a}, ``The contextual game experience: On the socio-cultural
  contexts for meaning in digital play,'' in \emph{Proc. DIGRA}, 2007, pp.
  810--814.

\bibitem{fay2013bootstrap}
N.~Fay, M.~Arbib, and S.~Garrod, ``How to bootstrap a human communication
  system,'' \emph{Cognitive science}, vol.~37, no.~7, pp. 1356--1367, 2013.

\bibitem{Groen20121575}
M.~Groen, M.~Ursu, S.~Michalakopoulos, M.~Falelakis, and E.~Gasparis,
  ``Improving video-mediated communication with orchestration,''
  \emph{Computers in Human Behavior}, vol.~28, no.~5, pp. 1575 -- 1579, 2012.

\bibitem{dake1995visual}
D.~M. Dake and B.~Roberts, ``The visual analysis of visual metaphor,'' 1995.

\bibitem{kievit2011semantic}
B.~Kievit-Kylar and M.~N. Jones, ``The semantic pictionary project,'' in
  \emph{Proc. Annual Conf. Cog. Sci. Soc.}, 2011, pp. 2229--2234.

\bibitem{saggar2015pictionary}
M.~Saggar \emph{et~al.}, ``Pictionary-based f{MRI} paradigm to study the neural
  correlates of spontaneous improvisation and figural creativity,''
  \emph{Nature (2005)}, 2015.

\bibitem{von2004labeling}
L.~Von~Ahn and L.~Dabbish, ``Labeling images with a computer game,'' in
  \emph{SIGCHI}.\hskip 1em plus 0.5em minus 0.4em\relax ACM, 2004, pp.
  319--326.

\bibitem{branson2010visual}
S.~Branson, C.~Wah, F.~Schroff, B.~Babenko, P.~Welinder, P.~Perona, and
  S.~Belongie, ``Visual recognition with humans in the loop,'' in
  \emph{European Conference on Computer Vision}.\hskip 1em plus 0.5em minus
  0.4em\relax Springer, 2010, pp. 438--451.

\bibitem{ullman2016atoms}
S.~Ullman, L.~Assif, E.~Fetaya, and D.~Harari, ``Atoms of recognition in human
  and computer vision,'' \emph{PNAS}, vol. 113, no.~10, pp. 2744--2749, 2016.

\bibitem{yu2015sketch}
Q.~Yu, Y.~Yang, Y.-Z. Song, T.~Xiang, and T.~M. Hospedales, ``Sketch-a-net that
  beats humans,'' \emph{arXiv preprint arXiv:1501.07873}, 2015.

\bibitem{seddati2015deepsketch}
O.~Seddati, S.~Dupont, and S.~Mahmoudi, ``Deepsketch: deep convolutional neural
  networks for sketch recognition and similarity search,'' in
  \emph{CBMI}.\hskip 1em plus 0.5em minus 0.4em\relax IEEE, 2015, pp. 1--6.

\bibitem{johnson2009games}
G.~Johnson and E.~Y.-L. Do, ``Games for sketch data collection,'' in
  \emph{Proceedings of the 6th eurographics symposium on sketch-based
  interfaces and modeling}.\hskip 1em plus 0.5em minus 0.4em\relax ACM, 2009,
  pp. 117--123.

\bibitem{donahue2015long}
J.~Donahue, L.~Anne~Hendricks, S.~Guadarrama, M.~Rohrbach, S.~Venugopalan,
  K.~Saenko, and T.~Darrell, ``Long-term recurrent convolutional networks for
  visual recognition and description,'' in \emph{CVPR}, 2015, pp. 2625--2634.

\bibitem{yeung2015every}
S.~Yeung, O.~Russakovsky, N.~Jin, M.~Andriluka, G.~Mori, and L.~Fei-Fei,
  ``Every moment counts: Dense detailed labeling of actions in complex
  videos,'' \emph{arXiv preprint arXiv:1507.05738}, 2015.

\bibitem{7299101}
J.~Y.-H. Ng, M.~Hausknecht, S.~Vijayanarasimhan, O.~Vinyals, R.~Monga, and
  G.~Toderici, ``Beyond short snippets: Deep networks for video
  classification,'' in \emph{CVPR}, 2015, pp. 4694--4702.

\bibitem{ma2016learning}
S.~Ma, L.~Sigal, and S.~Sclaroff, ``Learning activity progression in lstms for
  activity detection and early detection,'' in \emph{CVPR}, 2016, pp.
  1942--1950.

\bibitem{lev2016rnn}
G.~Lev, G.~Sadeh, B.~Klein, and L.~Wolf, ``Rnn fisher vectors for action
  recognition and image annotation,'' in \emph{ECCV}.\hskip 1em plus 0.5em
  minus 0.4em\relax Springer, 2016, pp. 833--850.

\bibitem{geman2015visual}
D.~Geman, S.~Geman, N.~Hallonquist, and L.~Younes, ``Visual turing test for
  computer vision systems,'' \emph{PNAS}, vol. 112, no.~12, pp. 3618--3623,
  2015.

\bibitem{malinowski2014towards}
M.~Malinowski and M.~Fritz, ``Towards a visual turing challenge,'' \emph{arXiv
  preprint arXiv:1410.8027}, 2014.

\bibitem{gao2015you}
H.~Gao, J.~Mao, J.~Zhou, Z.~Huang, L.~Wang, and W.~Xu, ``Are you talking to a
  machine? dataset and methods for multilingual image question,'' in
  \emph{NIPS}, 2015, pp. 2296--2304.

\bibitem{agnosia}
L.~Baugh, L.~Desanghere, and J.~Marotta, ``Agnosia,'' in \emph{Encyclopedia of
  Behavioral Neuroscience}.\hskip 1em plus 0.5em minus 0.4em\relax Academic
  Press, Elsevier Science, 2010, vol.~1, pp. 27--33.

\bibitem{sigurdsson2016hollywood}
G.~A. Sigurdsson, G.~Varol, X.~Wang, A.~Farhadi, I.~Laptev, and A.~Gupta,
  ``Hollywood in homes: Crowdsourcing data collection for activity
  understanding,'' in \emph{ECCV}, 2016.

\bibitem{eigen2015predicting}
D.~Eigen and R.~Fergus, ``Predicting depth, surface normals and semantic labels
  with a common multi-scale convolutional architecture,'' in \emph{Proceedings
  of the IEEE International Conference on Computer Vision}, 2015, pp.
  2650--2658.

\bibitem{DBLP:journals/corr/SaxeMG13}
\BIBentryALTinterwordspacing
A.~M. Saxe, J.~L. McClelland, and S.~Ganguli, ``Exact solutions to the
  nonlinear dynamics of learning in deep linear neural networks,'' \emph{CoRR},
  vol. abs/1312.6120, 2013. [Online]. Available:
  \url{http://arxiv.org/abs/1312.6120}
\BIBentrySTDinterwordspacing

\end{thebibliography}

\end{document}